\pdfoutput=1
\documentclass[11pt]{article}

\usepackage[final]{acl}

\usepackage{times}
\usepackage{latexsym}
\usepackage[T1]{fontenc}
\usepackage[utf8]{inputenc}
\usepackage{microtype}
\usepackage{inconsolata}
\usepackage{graphicx}

\usepackage{amsmath}
\usepackage{amssymb}
\usepackage{amsfonts}
\usepackage{mathtools}
\usepackage{amsthm}
\usepackage{dsfont}
\usepackage{nicefrac}

\usepackage[table,dvipsnames]{xcolor}
\usepackage{colortbl}

\usepackage{booktabs}
\usepackage{subcaption}
\usepackage{multirow}
\usepackage{makecell}
\usepackage{array}
\usepackage{adjustbox}
\usepackage{placeins}
\usepackage{cuted}
\usepackage{capt-of}
\usepackage{siunitx}
\usepackage{algorithm}
\usepackage{algpseudocode}

\usepackage{enumitem}
\usepackage{tcolorbox}
\usepackage{listings}
\tcbuselibrary{skins,breakable}

\usepackage[capitalize,noabbrev]{cleveref}

\usepackage{xurl}

\graphicspath{{figures/}}

\definecolor{bestbg}{HTML}{D1E8FF}
\definecolor{secondbg}{HTML}{EDF5FF}

\newcommand{\wb}[1]{\colorbox{white}{#1}}

\definecolor{bestc}{HTML}{D1E8FF}
\definecolor{secondc}{HTML}{EDF5FF}

\definecolor{tokbestc}{HTML}{D1E8FF}

\definecolor{deepblue}{RGB}{0,102,204}
\definecolor{deepgreen}{RGB}{0,153,102}
\definecolor{deepgray}{RGB}{120,120,120}

\definecolor{stepblue}{RGB}{48,129,208}
\definecolor{commentgreen}{RGB}{60,120,60}
\newcommand{\algphase}[1]{\State \textcolor{stepblue}{\textbf{#1}}}
\newcommand{\algcomm}[1]{\hfill \textcolor{commentgreen}{\footnotesize /* #1 */}}

\theoremstyle{plain}

\theoremstyle{definition}

\newcommand{\method}{\texttt{DeepLook}}

\lstdefinestyle{py}{
  language=Python,
  basicstyle=\ttfamily\scriptsize,
  breaklines=true,
  columns=fullflexible,
  showstringspaces=false,
  keywordstyle=\color{blue!60!black},
  commentstyle=\color{gray!70!black},
  stringstyle=\color{green!40!black},
  aboveskip=2pt,belowskip=2pt
}

\newtcolorbox{implbox}[2][]{%
  enhanced,
  breakable,
  colback=white,
  colframe=cyan!60!blue,
  arc=10pt,
  boxrule=0.9pt,
  left=8pt,right=8pt,top=6pt,bottom=6pt,
  title=\textbf{#2},
  coltitle=white,
  colbacktitle=cyan!70!blue,
  fonttitle=\normalsize,
  attach boxed title to top left={yshift=-2mm, xshift=1mm},
  boxed title style={arc=8pt, boxrule=0pt},
  #1
}

\newcommand{\restorepaperfloatspacing}{}

\title{\method{}: Uncertainty-Guided Branching with Lookahead Confidence for Efficient Reasoning}

\author{
  \textbf{Tingxin Yang\textsuperscript{1}}\thanks{Equal contribution.},
  \textbf{Zefeng Wang\textsuperscript{2}}\footnotemark[1],
  \textbf{Mengyue Wang\textsuperscript{1}},
  \textbf{Jinhe Bi\textsuperscript{2}},
\\
  \textbf{Xingcheng Zhou\textsuperscript{1}},
  \textbf{Volker Tresp\textsuperscript{2}},
  \textbf{Yunpu Ma\textsuperscript{2,3,4}}
\\
\\
  \textsuperscript{1}Technical University of Munich,
  \textsuperscript{2}LMU Munich,
  \textsuperscript{3}MCML,
  \textsuperscript{4}MemAgents Lab
\\
  \small{
    \textbf{Correspondence:} \href{mailto:cognitive.yunpu@gmail.com}{cognitive.yunpu@gmail.com}
  }
}

\begin{document}

\maketitle

\begin{abstract}
Test-time scaling improves large language model reasoning, but spends compute uniformly along a reasoning trace, even in stretches that are already stable.
Motivated by the observation that reasoning failures show an early onset of uncertainty, we introduce \method{}, a training-free decoding framework that monitors segment-level confidence, branches only at uncertainty bottlenecks, and ranks candidate continuations by \textit{Average Lookahead Confidence} (ALC)---their mean confidence over fixed-horizon rollouts---before completing and voting over a pruned subset.
On four competition mathematics benchmarks across four open models, \method{} matches or exceeds the accuracy of \texttt{DeepConf-low} in 12 of 16 settings while generating 87.3\% fewer tokens, showing that a model's own confidence is informative enough to decide \emph{where} extra decoding compute is worth spending.
On knowledge-intensive GPQA-Diamond the accuracy advantage disappears even though the token saving does not: coherent hallucinations also receive high ALC, so internal confidence localizes instability, not falsehood.
Code is available at \url{https://github.com/txyang1/deeplook}.
\end{abstract}

\begin{figure*}[t]
\centering
\includegraphics[width=0.95\textwidth]{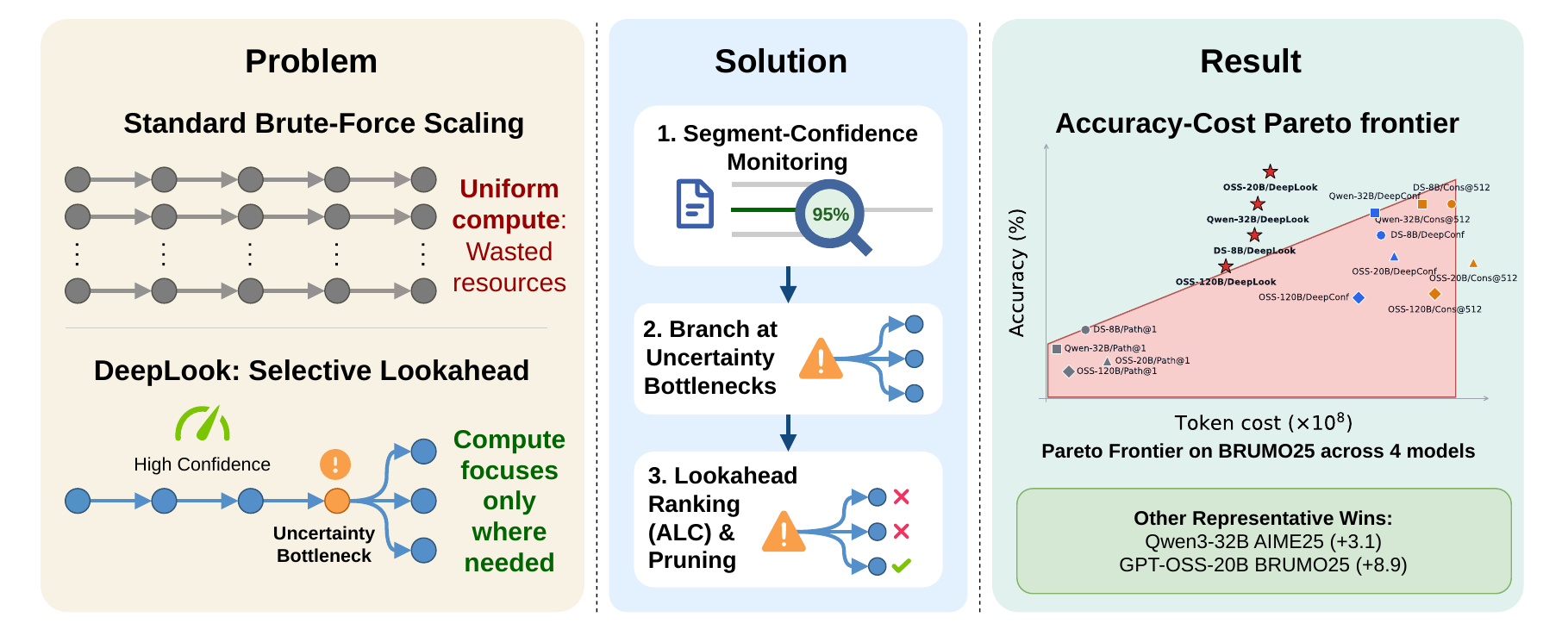}
\caption{\textbf{\method{} conceptual overview and accuracy--token-cost frontier.} Left: uniform brute-force scaling vs.\ selective lookahead at uncertainty bottlenecks. Middle: the \method{} pipeline of segment monitoring, branching, and ALC ranking. Right: \method{} on a higher-accuracy, lower-token-cost Pareto frontier, with an 87.3\% average generated-token reduction relative to \texttt{DeepConf-low} (details in \S\ref{sec:main_results}).}
\label{fig_teaser}
\end{figure*}

\section{Introduction}
\label{sec:intro}

Test-time scaling has become a central design axis for large language model (LLM) reasoning: instead of relying only on larger pretrained models, a growing family of methods allocates additional compute during decoding to sample, search, verify, or refine candidate solutions \citep{wei2022chain,wang2023selfconsistency,yao2023tree,lightman2024let,madaan2023selfrefine,snell2025scaling}. The key question is no longer whether extra inference-time compute helps, but how to spend it effectively when reasoning traces are long, error-prone, and expensive to generate.

\begin{figure}[t]
  \centering
  \begin{subfigure}{\columnwidth}
    \centering
    \includegraphics[width=0.88\linewidth]{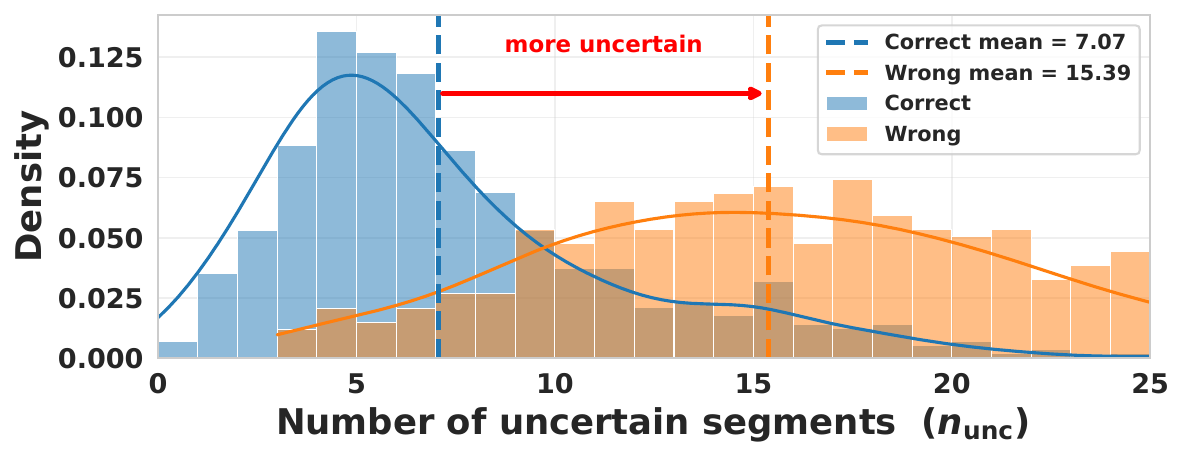}
    \caption{Count $n_{\mathrm{unc}}$.}
    \label{fig:kde-deepseek-nunc}
  \end{subfigure}

  \vspace{2pt}
  \begin{subfigure}{\columnwidth}
    \centering
    \includegraphics[width=0.88\linewidth]{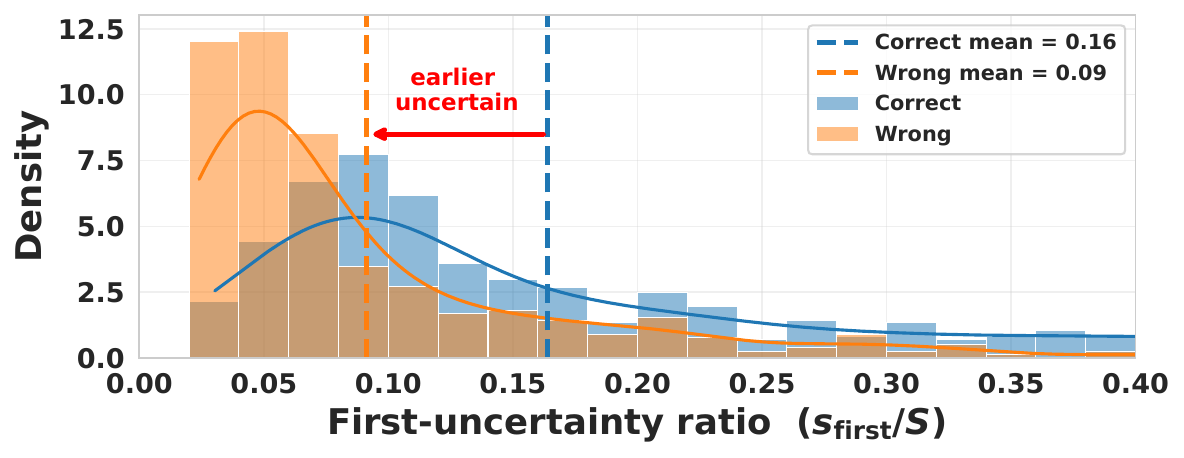}
    \caption{Onset ratio $r_{\mathrm{first}}$.}
    \label{fig:kde-deepseek-rfirst}
  \end{subfigure}
  \caption{\textbf{Uncertainty diagnostics correlate with errors on DeepSeek-R1-8B.} Wrong traces (orange) exhibit more uncertain segments (a) and an \textit{earlier uncertainty onset} (b) than correct traces (blue), motivating early confidence drops as triggers for lookahead exploration.}
  \label{fig:kde-deepseek-main}
\end{figure}

Current approaches leave this question only partly answered. Full-trajectory aggregation such as self-consistency regenerates long high-confidence prefixes that are largely shared across samples \citep{wang2023selfconsistency,brown2024llmmonkeys,snell2025scaling}. Confidence-aware methods such as DeepConf filter low-quality traces using model-internal signals \citep{fu2025deepconf}, yet these remain local or trace-level signals: without an explicit future check, an apparently plausible next step can still lead to a globally poor continuation \citep{ma2024nonmyopic,xu2025phidecoding}. Search and planning frameworks do evaluate alternatives beyond the next token, but their tree or graph expansion is ill-suited to intervening at a few uncertain points within an otherwise single reasoning trace \citep{yao2023tree,besta2024graph,zhou2024language}. This gap motivates our central question:

\begin{center}
\textbf{\textit{How can test-time reasoning compute be spent only where a trace becomes uncertain, without falling prey to the myopia of local confidence?}}
\end{center}

In response to this question, we conduct an empirical investigation into the model's internal uncertainty dynamics. 
As shown in Figure~\ref{fig:kde-deepseek-main}, correct and incorrect DeepSeek-R1-8B traces exhibit markedly different uncertainty profiles. Writing $n_{\mathrm{unc}}$ for the number of uncertain segments in a trace and $r_{\mathrm{first}}$ for the normalized position of the first one, incorrect traces contain many more uncertain segments on average (Figure~\ref{fig:kde-deepseek-nunc}) and encounter their first uncertain segment earlier in the generation (Figure~\ref{fig:kde-deepseek-rfirst}). Such drops mark where additional compute is likely to be most useful: rather than resampling complete trajectories, inference can intervene near persistent low-confidence bottlenecks. However, triggering alone only identifies \emph{when} to branch. It does not determine which continuation will remain reliable, motivating a future check before committing further compute.

Guided by this principle, we introduce \method{}, a confidence-triggered lookahead framework that first detects where reasoning becomes uncertain and then uses future confidence to select which continuation merits further computation. \method{} monitors segment-level confidence during decoding, triggers intervention only at local uncertainty bottlenecks, and ranks candidate continuations by \emph{Average Lookahead Confidence} (ALC), the mean segment-level confidence over fixed-horizon rollout continuations; selective triggering, lookahead evaluation, and adaptive pruning together preserve the robustness benefits of test-time scaling while sharply reducing wasted compute (Figure~\ref{fig_teaser}).

We evaluate \method{} on four competition-level benchmarks, AIME24, AIME25, BRUMO25, and HMMT25 \citep{jia_aime24_2024,matharena_aime25_2025,matharena_brumo25_2025,matharena_hmmt25_2025}, using four open-weight models: DeepSeek-R1-8B, Qwen3-32B, GPT-OSS-20B, and GPT-OSS-120B \citep{deepseek_r1,qwen3_techreport,openai_gptoss}. Across these settings, \method{} \textbf{shifts the accuracy--token-cost trade-off toward substantially lower inference cost}: it matches or exceeds the accuracy of \texttt{DeepConf-low} in 12 of 16 settings while generating on average \textbf{8.5$\times$} fewer tokens (5.1--16.0$\times$ across settings)---an average generated-token reduction of \textbf{87.3\%}. The advantage persists against much larger sampling budgets such as \texttt{Cons@512} (matched or exceeded in 13 of 16 settings). Targeted lookahead and pruning thus recover test-time exploration benefits without uniformly generating complete trajectories.

Our contributions are threefold:
\begin{itemize}[leftmargin=1.1em, labelsep=0.4em, itemsep=1pt, topsep=2pt, parsep=0pt, label=\textcolor{stepblue}{$\blacktriangleright$}]
  \item \textbf{Selective intervention.} \method{}, a training-free framework that branches only at persistent uncertainty bottlenecks, leaving stable stretches of a trace untouched.
  \item \textbf{Non-myopic ranking.} \textit{Average Lookahead Confidence} (ALC), which ranks continuations by their mean confidence over fixed-horizon rollouts rather than by local confidence alone.
  \item \textbf{Accuracy--cost evidence.} An evaluation over four benchmarks and four models, plus a boundary analysis on knowledge-intensive GPQA-Diamond, where internal-confidence intervention stops helping.
\end{itemize}

\section{Related work}
\label{sec:related}

\paragraph{Efficient test-time scaling.}
Inference-time reasoning improves LLM reliability through chain-of-thought (CoT) prompting~\citep{wei2022chain, kojima2022zeroshot}, repeated sampling~\citep{wang2023selfconsistency, brown2024llmmonkeys}, and compound inference systems~\citep{chen2024morecalls}, but these gains often come with large token costs~\citep{feng2025efficientsurvey, sui2025stopoverthinking, qu2025efficientreasoningsurvey}. Recent work spends test-time compute more economically via adaptive compute control~\citep{aggarwal2025l1, zhang2025adaptthink, alkhalili2026forkthink, scalena2026eager}, adaptive sample budgets for self-consistency~\citep{aggarwal2023adaptiveconsistency, li2024esc}, reasoning compression~\citep{xia2025tokenskip, xu2025chainofdraft}, pruning~\citep{hou2025thinkprune}, verifier-guided inference~\citep{lightman2024let}, and feedback-based refinement~\citep{madaan2023selfrefine}. Yet many effective strategies still operate over complete trajectories, regenerating long shared prefixes for traces that differ only around a few pivotal steps. \method{} targets this redundancy by moving from full-trajectory scaling to selective completion, spending extra compute only where an online signal flags a segment worth revisiting.

\paragraph{Confidence-aware computation.} 
Model-internal and online generation signals provide a natural way to decide \emph{when} extra inference should be spent, supporting confidence-based early exit, partial-trace pruning~\citep{yang2025earlyexit,tu2025deepprune}, answer selection, and trace filtering~\citep{kang2025selfcertainty, fu2025deepconf}. Beyond broad evidence that model probabilities, semantic uncertainty, and self-evaluation track generation quality~\citep{kadavath2022language, kuhn2023semantic}, this uncertainty is concentrated rather than diffuse: a small minority of high-entropy tokens act as forks that steer the trace toward different reasoning paths~\citep{wang2025forkingtokens}, and resampling at those specific positions---but not others---changes the outcome~\citep{bigelow2025forkingpaths}. This concentration is what makes a selective, trigger-driven intervention viable. Closest in signal is DEER~\citep{yang2025earlyexit}, which monitors similar confidence dynamics but makes the \emph{opposite}, complementary intervention: it exits reasoning early when the model becomes confident, whereas \method{} branches where confidence drops. Because token-level probabilities are noisy and whole-trace scores too coarse to locate where a solution destabilizes, \method{} uses confidence primarily as a \emph{trigger}: persistent local uncertainty determines where to branch, and a separate lookahead signal evaluates which continuations should be kept.

\paragraph{Exploration in sampling.}
Exploration-based inference addresses the complementary question of \emph{how} to evaluate alternatives once branching is allowed. Work on non-myopic decoding directly targets the limits of next-token likelihood by using short-horizon foresight to score candidates by their downstream behavior~\citep{ma2024nonmyopic, xu2025phidecoding}; \citet{Bachmann2024ThePO} trace the same difficulty to teacher-forced training itself. Search-based reasoning methods instead expand larger spaces of thoughts, trajectories, or rollouts~\citep{yao2023tree, mits2025}, an idea that recurs in RL training~\citep{latr2025lookahead}; related planning and agentic frameworks explore thought graphs, simulated states, and action trajectories~\citep{besta2024graph, hao2023reasoning, zhou2024language, yao2023react}, while latent deliberation internalizes future-aware computation before visible generation~\citep{zelikman2024quietstar}. Likelihood-guided resampling has also been applied globally: \citet{karan2025reasoningsampling} sharpen entire traces with an MCMC-style procedure driven by the model's own likelihoods, whereas \method{} applies confidence \emph{locally}, deciding where within a single trace to branch and what to keep. These methods motivate future-sensitive evaluation, but broad search can introduce substantial branching overhead; \method{} adopts the foresight principle in a narrower form, performing fixed-budget lookahead only at confidence-triggered uncertainty bottlenecks, without expanding a full search tree.

\paragraph{Process reward models and step verifiers.}
A complementary line trains process reward models on step-level supervision to score intermediate reasoning~\citep{lightman2024let} and uses these verifiers to guide beam search or best-of-$N$ selection; \citet{xie2023selfevaluation} instead take the model's own stepwise self-evaluation as the guidance signal within stochastic beam search. Such verifiers offer strong signals but require task-specific annotation and additional trained components. \method{}'s ranking signal is derived entirely from the model's own confidence distributions---no supervision, no extra models---at the cost of being blind to confidently stated factual errors, a boundary we characterize empirically in \S\ref{sec:gpqa_boundary}.

\begin{figure*}[t]
  \centering
  \includegraphics[width=0.95\textwidth]{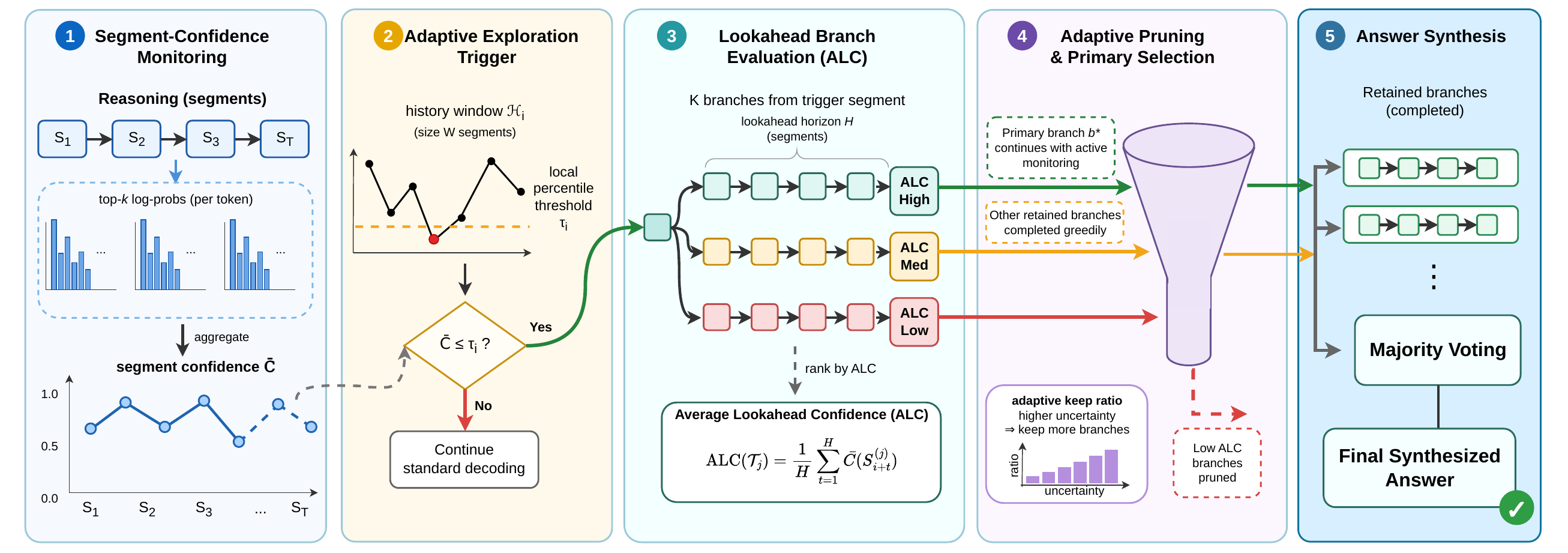}
  \caption{
  \textbf{\method{} pipeline.}
  (1) Token-level top-$k$ log-probabilities are aggregated into a segment confidence $\bar{C}(S_i)$, which (2) is compared against a percentile threshold $\tau_i$ over a local history window $\mathcal{H}_i$, so exploration fires only on a drop relative to the trace's own recent history.
  (3) At a trigger, $K$ branches are rolled forward over a horizon $H$ and scored by Average Lookahead Confidence (ALC).
  (4) Branches are ranked and adaptively pruned, the best branch $b^*$ resuming as the primary trajectory while the other retained branches are completed.
  (5) All completed traces are majority-voted into the final answer.
  }
  \label{fig:first_fig}
\end{figure*}

\section{\method{}}
\label{sec:method}

\method{} is a monitor-and-intervene decoding framework organized around the five stages shown in Figure~\ref{fig:first_fig} and summarized in Algorithm~\ref{alg:deeplook_compact}. It converts token-level top-$k$ log-probability statistics into segment-level confidence and intervenes only when the current segment becomes uncertain relative to its recent history. At an intervention point, \method{} performs lookahead over $K$ candidate continuations, ranks them by Average Lookahead Confidence (ALC), adaptively retains a subset according to the observed uncertainty, and synthesizes the final answer from the completed retained traces.

\subsection{Confidence in reasoning and segment-level triggering}
\label{sec:confidence_trigger}

\method{} monitors confidence at the segment level to decide when additional computation is needed. Following DeepConf \citep{fu2025deepconf}, we define token confidence $C_t$ from the top-$k$ next-token distribution:
\begin{equation}
\label{eq:token_conf_min}
C_t \;=\; -\frac{1}{k}\sum_{\ell=1}^k \log p_t\!\left(v_t^{(\ell)}\right),
\end{equation}
where $v_t^{(1)},\dots,v_t^{(k)}$ are the top-$k$ tokens at decoding step $t$.

\begin{algorithm}[t]
\caption{\method{} (compact). Full pseudocode with all bookkeeping: Appendix Algorithm~\ref{alg:deeplook_overview_min}.}
\label{alg:deeplook_compact}
\small
\begin{algorithmic}[1]
\State \textbf{Input:} prompt $x$; width $K$; horizon $H$; rounds $R_{\max}$
\State $P \gets x$;\; completed traces $\mathcal{T} \gets \emptyset$;\; $r \gets 0$
\While{no final answer}
    \State decode segment $S$; score $\bar{C}(S)$, $\tau$ \algcomm{Eqs.~\eqref{eq:token_conf_min}--\eqref{eq:adaptive_threshold}}
    \If{uncertain per Eq.~\eqref{eq:explore_trigger} \textbf{and} $r < R_{\max}$}
        \State $r \gets r+1$; discard $S$; sample $K$ branches
        \State roll $H$ segments ahead; rank by ALC \algcomm{Eq.~\eqref{eq:alc}}
        \State keep top $\lfloor K\rho_i \rfloor$ branches \algcomm{Eq.~\eqref{eq:dyn_keep_simple}}
        \State best branch $b^*$ resumes as main path: $P \gets P \oplus b^*$
        \State complete other kept branches; add to $\mathcal{T}$
    \Else
        \State accept segment: $P \gets P \oplus S$
    \EndIf
\EndWhile
\State \Return majority vote over $\mathcal{T} \cup \{P\}$ \algcomm{Eq.~\eqref{eq:majority_vote}}
\end{algorithmic}
\end{algorithm}

However, token-level metrics are inherently volatile: individual token probabilities fluctuate due to local lexical ambiguity even when the underlying logic is correct (visualized in Appendix Figure~\ref{fig:combined_conf_vis}). To reduce this noise, \method{} decodes the main trajectory in fixed-length segments and aggregates token confidence over each segment $S_i$ of length $L_{\mathrm{main}}$:

\begin{equation}
\label{eq:seg_conf_min}
\bar{C}(S_i) \;=\; \frac{1}{L_{\mathrm{main}}}\sum_{t \in S_i} C_t.
\end{equation}

Rather than use a fixed confidence cutoff, \method{} compares the current segment against a local history window of recent segment confidences. This local comparison makes the trigger depend on relative confidence changes within the current trace, reducing sensitivity to the absolute calibration of different models. For segment $S_i$, we define
\begin{equation}
\label{eq:adaptive_threshold}
\begin{aligned}
\tau_i &= \text{Percentile}(\mathcal{H}_i, q), \\
\text{where } \mathcal{H}_i &= \{ \bar{C}(S_{j}) \}_{j=i-W+1}^{i}
\end{aligned}
\end{equation}
is the sliding window of the most recent $W$ segment confidences (including the current segment); triggering is disabled during a short warmup while the window first fills. Lookahead exploration is governed by an uncertainty state with hysteresis:
\begin{equation}
\label{eq:explore_trigger}
\text{enter: } \bar{C}(S_i) \leq \tau_i, \qquad \text{exit: } \bar{C}(S_i) \geq (1+\delta)\,\tau_i.
\end{equation}
Entering the uncertain state triggers exploration, and the state persists across subsequent segments---each further segment triggering again while the branch budget lasts---until confidence recovers above the release bound; the margin $\delta$ prevents oscillating re-triggers from minor fluctuations near the threshold, focusing branching on local instability rather than on uniformly difficult regions.
The design operationalizes the early-onset observation of Figure~\ref{fig:kde-deepseek-main} directly: what distinguishes incorrect traces is not their absolute confidence level, which varies across models and problems, but the arrival of segments markedly less confident than the trace's own recent history. This is exactly the event the relative percentile test detects, and the earlier it fires, the more of the trace remains correctable. We treat this as a correlational signal for allocating compute, not a causal account of why reasoning fails; \S\ref{sec:ablation_trigger_signal} reports the correct-vs-incorrect separation of the trigger statistic under all component variants.
In our experiments, we use a single fixed configuration ($q=0.10$, $W=8$, $\delta=0.02$) across all four model families without model-specific recalibration.

\subsection{Lookahead exploration and Average Lookahead Confidence}
\label{sec:lookahead_exploration}

Once a confidence drop triggers exploration, \method{} suspends main-path decoding and compares several continuations before committing further computation. Because a continuation can have reasonable immediate probability yet lead to an unstable subsequent trace, \method{} follows the non-myopic decoding principle~\citep{ma2024nonmyopic} and evaluates candidates by their near-future behavior rather than by the trigger segment alone.

Specifically, \method{} discards the trigger segment $S_i$ and spawns $K$ parallel candidate branches $\{b_1,\dots,b_K\}$ from the prefix preceding it, using seeded nucleus sampling to encourage diversity among local continuations.
Each branch $b_j$ first generates a rollout segment of length $L_{\mathrm{roll}}$---the text that replaces $S_i$ if the branch is adopted---and is then extended for a lookahead horizon of $H$ further segments of length $L_{\mathrm{look}}$, which are used only for scoring and then discarded.
We score a branch by Average Lookahead Confidence (ALC), the mean segment confidence observed over this lookahead window:
\begin{equation}
\label{eq:alc}
\mathrm{ALC}(b_j) = \frac{1}{H}\sum_{h=1}^{H} \bar{C}(S_{h}^{(j)}),
\end{equation}
where $\bar{C}(S_{h}^{(j)})$ denotes the confidence of the $h$-th lookahead segment of branch $b_j$.
ALC is an internal stability heuristic, not an external correctness verifier: branches that return to confident generation score higher, a future-sensitive ranking that avoids committing to a locally plausible but unstable continuation. Branches are ranked by descending ALC and passed to the pruning stage.

\begin{table*}[t]

  \centering
  \begin{scriptsize}
  \setlength{\aboverulesep}{0.5pt}     %
  \setlength{\belowrulesep}{0.5pt}
  \setlength{\tabcolsep}{5pt}
  \renewcommand{\arraystretch}{0.7}

  \resizebox{\textwidth}{!}{
  \begin{tabular}{llccccc|cccc}
    \toprule
    \multirow{2}{*}{Model} & \multirow{2}{*}{Dataset} & \multirow{2}{*}{Path@1} &
    \multicolumn{2}{c}{Cons@512} &
    \multicolumn{2}{c}{DeepConf-low} &
    \multicolumn{4}{c}{\method{}@128 (Ours)} \\
    \cmidrule(lr){4-5}\cmidrule(lr){6-7}\cmidrule(lr){8-11}
      & & &
    Tok & Acc &
    Tok & Acc &
    Tok & $\Delta\%\downarrow$ (C\,/\,DC) & Acc & $\Delta$ (C\,/\,DC) \\
    \midrule

    \multirow{4}{*}{DeepSeek-R1-8B}
      & AIME24  & 83.0\% & 3.55 & 86.7\% & 0.78 & \colorbox{secondbg}{92.5\%} & 0.121 & (-96.6\%/-84.5\%) & \colorbox{bestbg}{93.3\%} & (+6.6/+0.8) \\
      & AIME25  & 76.9\% & 4.01 & 82.3\% & 1.24 & \colorbox{secondbg}{86.4\%} & 0.127 & (-96.8\%/-89.8\%) & \colorbox{bestbg}{86.7\%} & (+4.4/+0.3) \\
      & BRUMO25 & 80.0\% & 3.56 & \colorbox{bestbg}{93.3\%} & 1.07 & \colorbox{secondbg}{90.0\%} & 0.124 & (-96.5\%/-88.4\%) & \colorbox{secondbg}{90.0\%} & (-3.3/+0.0) \\
      & HMMT25  & 58.1\% & 4.49 & 69.8\% & 1.60 & \colorbox{bestbg}{77.6\%} & 0.213 & (-95.3\%/-86.7\%) & \colorbox{secondbg}{73.3\%} & (+3.5/-4.3) \\
    \midrule

    \multirow{4}{*}{Qwen3-32B}
      & AIME24  & 80.6\% & 2.00 & 84.8\% & 0.66 & \colorbox{secondbg}{89.5\%} & 0.102 & (-94.9\%/-84.5\%) & \colorbox{bestbg}{90.0\%} & (+5.2/+0.5) \\
      & AIME25  & 71.7\% & 2.43 & 80.1\% & 1.14 & \colorbox{secondbg}{80.2\%} & 0.140 & (-94.2\%/-87.7\%) & \colorbox{bestbg}{83.3\%} & (+3.2/+3.1) \\
      & BRUMO25 & 78.0\% & 2.17 & \colorbox{bestbg}{93.3\%} & 0.96 & \colorbox{secondbg}{92.4\%} & 0.131 & (-94.0\%/-86.4\%) & \colorbox{bestbg}{93.3\%} & (+0.0/+0.9) \\
      & HMMT25  & 51.9\% & 2.76 & \colorbox{secondbg}{63.4\%} & 1.55 & \colorbox{bestbg}{64.5\%} & 0.097 & (-96.5\%/-93.7\%) & \wb{63.3\%} & (-0.1/-1.2) \\
    \midrule

    \multirow{4}{*}{GPT-OSS-20B}
      & AIME24  & 92.1\% & 5.57 & \colorbox{bestbg}{96.7\%} & 1.11 & \colorbox{secondbg}{95.7\%} & 0.168 & (-97.0\%/-84.9\%) & \colorbox{bestbg}{96.7\%} & (+0.0/+1.0) \\
      & AIME25  & 91.7\% & 6.26 & 95.4\% & 1.21 & \colorbox{secondbg}{96.1\%} & 0.235 & (-96.2\%/-80.6\%) & \colorbox{bestbg}{96.7\%} & (+1.3/+0.6) \\
      & BRUMO25 & 76.7\% & 5.16 & 87.1\% & 1.34 & \colorbox{secondbg}{87.8\%} & 0.162 & (-96.9\%/-87.9\%) & \colorbox{bestbg}{96.7\%} & (+9.6/+8.9) \\
      & HMMT25  & 86.7\% & 8.16 & \colorbox{secondbg}{89.9\%} & 2.17 & 89.4\% & 0.189 & (-97.7\%/-91.3\%) & \colorbox{bestbg}{90.0\%} & (+0.1/+0.6) \\
    \midrule

    \multirow{4}{*}{GPT-OSS-120B}
      & AIME24  & 91.9\% & 2.66 & \colorbox{secondbg}{96.7\%} & 0.53 & \colorbox{bestbg}{97.0\%} & 0.079 & (-97.0\%/-85.1\%) & \colorbox{secondbg}{96.7\%} & (+0.0/-0.3) \\
      & AIME25  & 91.8\% & 3.23 & \colorbox{secondbg}{97.1\%} & 0.49 & \colorbox{bestbg}{97.9\%} & 0.072 & (-97.8\%/-85.3\%) & \wb{96.7\%} & (-0.4/-1.2) \\
      & BRUMO25 & 75.6\% & 2.68 & \colorbox{secondbg}{83.8\%} & 0.73 & 83.4\% & 0.076 & (-97.2\%/-89.6\%) & \colorbox{bestbg}{86.7\%} & (+2.9/+3.3) \\
      & HMMT25  & 78.9\% & 4.09 & \colorbox{secondbg}{92.8\%} & 0.97 & 92.0\% & 0.094 & (-97.7\%/-90.3\%) & \colorbox{bestbg}{93.3\%} & (+0.5/+1.3) \\
    \bottomrule
  \end{tabular}}
  \end{scriptsize}
  \caption{\textbf{Main results.} Accuracy (\%) and total generated-token cost (Tok, $\times 10^8$, all inference stages). For \method{}@128, $\Delta\%\downarrow$ and $\Delta$ give token and accuracy change vs.\ (C\,/\,DC) $=$ (Cons@512\,/\,DeepConf-low). Shading marks the \colorbox{bestbg}{best} and \colorbox{secondbg}{second-best} accuracy per row. All accuracies are measured on $N{=}30$ problems per benchmark.}
  \label{tab:main_results}
\end{table*}

\subsection{Lookahead adaptive pruning and final answer synthesis}
\label{sec:adaptive_pruning}

After ALC ranking, \method{} keeps only a subset of branches for completion. Let $\Delta_i=\max(0,\tau_i - \bar{C}(S_i))$ denote the confidence gap at the trigger point. The keep ratio $\rho_i$ increases with this gap so that stronger uncertainty preserves more candidate answers:
\begin{equation}
\label{eq:dyn_keep_simple}
\begin{aligned}
\rho_u &= \rho_{\min} + \text{clip}\!\left(\frac{\Delta_i}{s}, 0, 1\right)(\rho_{\max} - \rho_{\min}), \\
\rho_i &= \text{clip}\!\left(\frac{\rho_u + \rho_{\text{base}}}{2},\; \rho_{\min},\; \rho_{\max}\right),
\end{aligned}
\end{equation}
where $[\rho_{\min},\rho_{\max}]$ bounds the keep ratio, $s$ controls sensitivity, and $\rho_{\text{base}}$ is the default retention rate; averaging with $\rho_{\text{base}}$ damps the gap-driven term, and when $\Delta_i = 0$ (possible while the trigger remains latched) the keep ratio falls back to $\rho_{\min}$.

Among branches satisfying $\mathrm{ALC}(b) \geq \tau_i$ and $\mathrm{ALC}(b) > \bar{C}(S_i)$, \method{} selects the highest-ALC branch as the primary trajectory $b^*$ (falling back to the highest rollout confidence---the mean token confidence of a branch's rollout segment---if none qualifies). The primary trajectory resumes segmented decoding under active monitoring and remains eligible for future interventions; the other retained branches---the top-$\lfloor K\rho_i \rfloor$ ALC-ranked candidates (at least one; ties broken by rollout confidence), minus $b^*$ when it ranks among them---are completed with the main-path sampling parameters without recursive triggering, and the rest are discarded.

Finally, \method{} extracts an answer $a_r$ from the primary trajectory and from each completed retained trace, skipping traces without a parseable boxed answer, and returns the majority vote over the resulting pool $\mathcal{T}$:
\begin{equation}
\label{eq:majority_vote}
\hat{a} = \operatorname*{arg\,max}_{a} \sum_{r \in \mathcal{T}} \mathbb{I}(a_r = a).
\end{equation}

Ties are resolved by the average confidence of the traces supporting each answer, keeping the voting rule standard and lightweight.

\section{Experiments}
\label{sec:experiments}

\subsection{Experimental settings}
\label{sec:settings}

\textbf{Benchmarks and metrics.}
We evaluate \method{} on four competition-level mathematics benchmarks: AIME24~\citep{jia_aime24_2024}, AIME25~\citep{matharena_aime25_2025}, BRUMO25~\citep{matharena_brumo25_2025}, and HMMT25~\citep{matharena_hmmt25_2025}, reporting \textbf{Accuracy (Acc)} of the final synthesized answer and \textbf{Token cost (Tok)}, the dataset-level total of generated tokens over all inference stages (main-path decoding, branch and lookahead rollouts, retained-branch completions), plus relative reduction ($\Delta\%$). Token counts measure generated-token work only; they exclude wall-clock latency and the KV-cache memory needed to evaluate $K$ branches in parallel (Appendix~\ref{sec:token_breakdown}). Separately from this accuracy--cost evaluation, we probe the knowledge-intensive regime on GPQA-Diamond~\citep{rein2023gpqa} in \S\ref{sec:gpqa_boundary}.

\textbf{Models.}
We use four open-weight models spanning 8B--120B: DeepSeek-R1-8B\footnote{The Qwen3-8B model distilled from DeepSeek-R1 (0528): \url{https://huggingface.co/deepseek-ai/DeepSeek-R1-0528-Qwen3-8B}.}, a distilled reasoning model; Qwen3-32B~\citep{qwen3_techreport}, a dense model; and GPT-OSS-20B/120B~\citep{openai_gptoss}.

\textbf{Baselines.}\label{sec:baselines}
We write \method{}@$B$ for a total branch budget of $B{=}K\times R_{\max}$ candidate continuations per problem; \method{}@128 uses $K{=}64$ branches over $R_{\max}{=}2$ rounds (Appendix~\ref{sec:appendix-gen-hparams}).
We compare against \textbf{(1)} \texttt{Path@1}, a single sampled trajectory; \textbf{(2)} \texttt{Cons@512}, computation-heavy self-consistency (512 samples; temperature 0.6, top-$p$ 0.95, matching \method{}'s rollout parameters) as a brute-force ceiling; and \textbf{(3)} \texttt{DeepConf-low}~\citep{fu2025deepconf}, an adaptive confidence-guided method tuned for efficiency, our primary resource-allocation baseline.

\subsection{Main results}
\label{sec:main_results}

Table~\ref{tab:main_results} reports accuracy and token cost across model scales; Appendix Figure~\ref{fig:main_results_pareto_by_dataset} plots the corresponding trade-off per benchmark.

\textbf{Surgical precision over brute force.}
Relative to \texttt{DeepConf-low}, \method{}@128 matches or exceeds accuracy in 12 of 16 settings (11 wins, 1 tie, 4 losses) while generating \textbf{87.3\%} fewer tokens (a mean per-setting ratio of \textbf{8.5$\times$}); on BRUMO25 with GPT-OSS-20B it gains \textbf{+8.9 points} (87.8\% $\to$ 96.7\%) at about \textbf{12\%} of the baseline's tokens. A two-sided sign test over the 15 non-tied settings (11 wins vs.\ 4 losses) does not distinguish the two methods on accuracy ($p{=}0.118$), whereas the token reduction holds in \textbf{16 of 16} settings (80.6\%--93.7\%, $p{<}10^{-4}$). The result we defend is therefore an order-of-magnitude cheaper decode at statistically indistinguishable accuracy. Table~\ref{tab:main_results} also shows where that saving costs something: on HMMT25 with DeepSeek-R1-8B and Qwen3-32B a modest accuracy drop accompanies the compute reduction, shrinking as the branch budget grows (66.7\% at @32 vs.\ 73.3\% at @128 on DeepSeek-R1-8B; Appendix Table~\ref{tab:main_results_full}), consistent with a completion-budget limitation on especially difficult instances.

\textbf{The branch budget is a well-behaved dial.} Accuracy is non-decreasing in the budget in all 16 settings (@32 $\le$ @64 $\le$ @128), with @128 strictly above @32 in 15 of them ($p{<}10^{-4}$, two-sided sign test), so \method{} exposes a monotone accuracy--cost knob rather than a single tuned operating point. The cheaper settings trade accuracy for cost as expected: @32 uses $32.6\times$ fewer tokens than \texttt{DeepConf-low} at $-3.9$ points on average, @64 $15.8\times$ at $-1.6$. We report @128 as the main configuration because it is the cheapest budget that reaches the baseline's accuracy level; Appendix Table~\ref{tab:main_results_full} gives the full curve.

\textbf{Outperforming the compute ceiling.}
The comparison with \texttt{Cons@512} highlights the diminishing returns of blind test-time scaling: on AIME25 with Qwen3-32B, \method{} improves over \texttt{Cons@512} by \textbf{+3.2 points} (80.1\% $\to$ 83.3\%) at roughly \textbf{17$\times$ less compute}. Figure~\ref{fig:hmmt_short_compressed} summarizes one such case: the consensus baseline repeatedly commits to a locally plausible heuristic, while branching at detected instability recovers the correct answer---consistent with the benefit coming from intervening where the trace becomes unstable rather than from candidate count (\S\ref{sec:ablation_component}).

\begin{figure}[t]
    \centering
    \footnotesize
    \tcbset{
        boxsep=1pt,
        left=2pt, right=2pt, top=1pt, bottom=1pt,
        toptitle=1pt, bottomtitle=1pt,
        arc=0.5mm, boxrule=0.5pt
    }

    \begin{tcolorbox}[colback=gray!2, colframe=gray!20]
        \textbf{Problem (HMMT'25):} Maximize door traversals in $11 \times 11$ grid loop (Edges=220; Odd V=36).
    \end{tcolorbox}

    \begin{tcolorbox}[colback=orange!3, colframe=orange!15, title=\textbf{\textcolor{orange!70!black}{Cons@512 $\rightarrow$ Incorrect (\boxed{202})}}]
            \textbf{Blind Heuristic:} $\text{Remove} = \text{Odd}/2 = 18$ \\[-0.6ex]
            \textbf{Failure:} Assumes local pairings always possible. \\[-0.6ex]
            \textbf{Final:} $220 - 18 = \mathbf{202}$ (Topologically impossible).
    \end{tcolorbox}

    \begin{tcolorbox}[colback=cyan!3, colframe=cyan!15, title=\textbf{\textcolor{cyan!70!black}{\method{} $\rightarrow$ Correct (\boxed{200})}}]
            \textbf{Lookahead Signal:} Detects boundary parity mismatch. \\[-0.6ex]
            \textbf{Correction:} Forced corner pairing adds +2 cost. \\[-0.6ex]
            \textbf{Outcome:} $220 - (18 + 2) = \mathbf{200}$.
    \end{tcolorbox}

    \caption{\textbf{Illustrative case summary (not a verbatim trace).} Hand-condensed decisive steps of an HMMT25 problem, contrasting the consensus failure mode of \texttt{Cons@512} with the \method{} run that recovered the correct answer; the summarization convention is described in Appendix~\ref{sec:appendix-examples}.}
    \label{fig:hmmt_short_compressed}
\end{figure}

\begin{figure}[t]
  \centering
  \includegraphics[width=0.95\columnwidth]{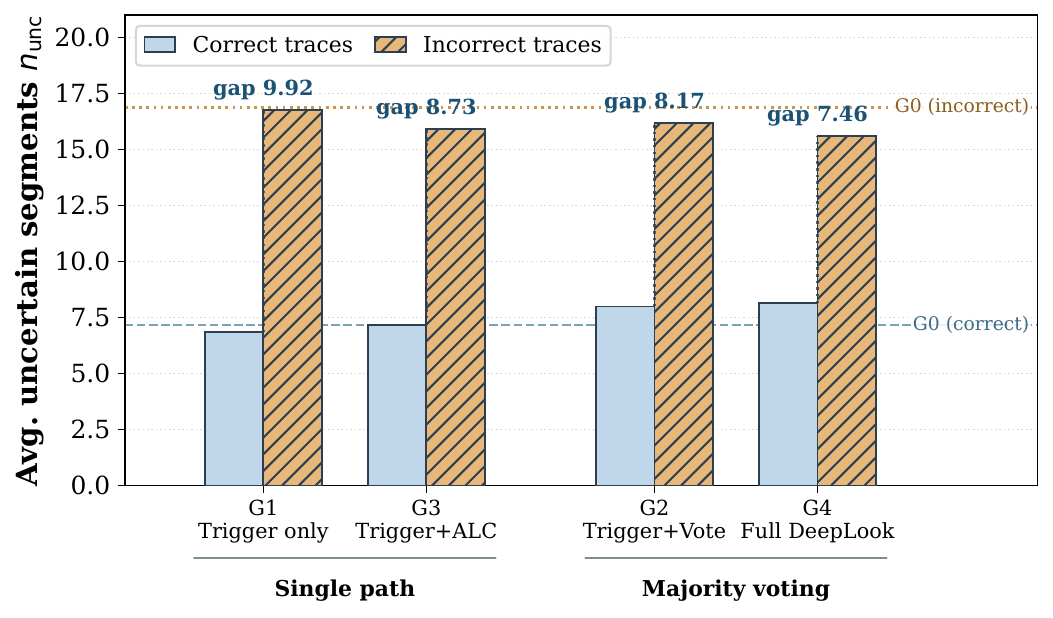}
  \caption{\textbf{Average $n_{\text{unc}}$ for correct vs.\ incorrect traces}, DeepSeek-R1-8B on HMMT25, across four component variants under both readout regimes; G0 (greedy decoding) is shown as the two reference lines. The correct--incorrect gap ranges over $7.46$--$9.92$ segments and never closes.}
  \label{fig:uncertainty-deepseek}
\end{figure}

\begin{figure*}[t]
    \centering
    \begin{subfigure}{0.41\textwidth}
        \centering
        \includegraphics[width=\linewidth]{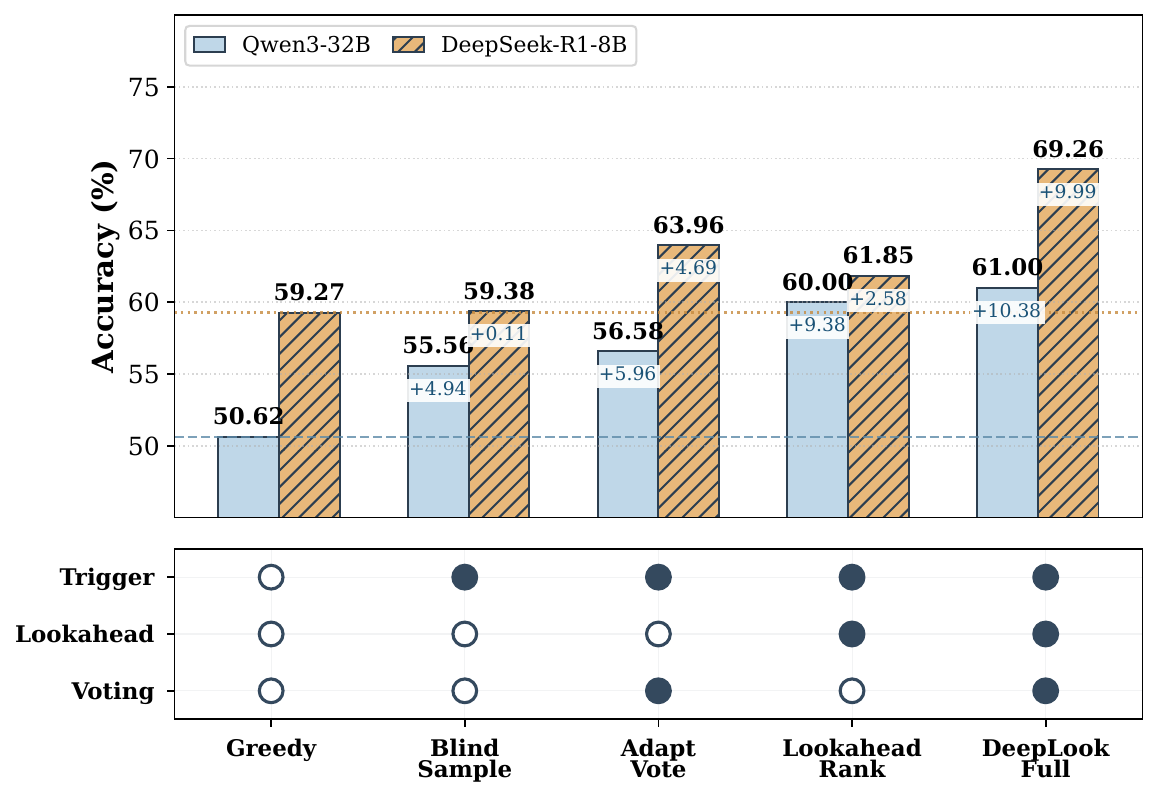}
        \caption{Component ablation on HMMT25.}
        \label{fig:ablation_sub}
    \end{subfigure}
    \hfill
    \begin{subfigure}{0.47\textwidth}
        \centering
        \includegraphics[width=\linewidth]{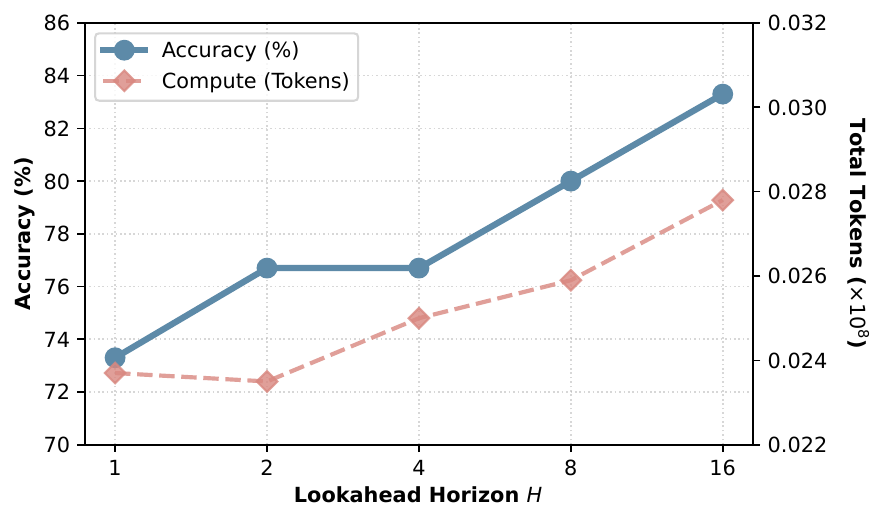}
        \caption{Impact of lookahead horizon $H$ on AIME25.}
        \label{fig:scaling_sub}
    \end{subfigure}

    \caption{\textbf{Ablation and scaling analysis.} (a) Accuracy as components are integrated (variants G0--G4 of \S\ref{sec:ablation_trigger_signal}); $\Delta$ is the gain over Greedy. (b) Accuracy as a function of the lookahead horizon $H$.}
    \label{fig:main_results}
\end{figure*}

\subsection{Does the trigger signal track reasoning fragility?}
\label{sec:ablation_trigger_signal}

\S\ref{sec:main_results} shows that acting on segment-level confidence pays off. Whether the gain comes from the signal itself or merely from the extra sampling it triggers is a separate question; two checks address it.

\textbf{The correct--incorrect separation survives every component configuration.}
Figure~\ref{fig:uncertainty-deepseek} reports the mean uncertain-segment count $n_{\text{unc}}$ for correct and incorrect traces across the greedy baseline G0 (\emph{Greedy}) and four component variants---G1 (\emph{Blind-Sample}: trigger, random branch), G2 (\emph{Adapt-Vote}: trigger + vote, no ranking), G3 (\emph{Lookahead-Rank}: trigger + ALC ranking), and G4 (full \method{})---on HMMT25 with DeepSeek-R1-8B, under both single-path and majority-voting readouts. The gap never closes: incorrect traces carry between $7.46$ and $9.92$ more uncertain segments than correct ones, in every configuration. These variants differ precisely in what happens \emph{after} triggering---whether candidates are ranked by ALC, whether outputs are voted---so a separation that is stable across all of them is a property of the trigger statistic itself, not of the machinery built on top of it.

\textbf{The pattern is not specific to one model family.}
The diagnostics that motivate \method{} in Figure~\ref{fig:kde-deepseek-main} replicate on Qwen3-32B: incorrect traces contain roughly twice as many uncertain segments ($11.60$ vs.\ $5.68$) and an earlier first uncertain segment (position ratio $0.24$ vs.\ $0.41$; Appendix~\ref{sec:appendix-qwen-kde}).

Together the two checks supply the method's premise: $n_{\text{unc}}$ is a correlate of trace-level correctness, stable across both model families examined, making \emph{where to intervene} decidable at decoding time---though not a correctness certificate for any individual continuation, a distinction that matters on knowledge-intensive tasks (\S\ref{sec:gpqa_boundary}).

\subsection{Component ablation} \label{sec:ablation_component}

Figure~\ref{fig:ablation_sub} isolates the three components---triggering, ALC ranking, and voting---on HMMT25, over the configurations G0--G4 of \S\ref{sec:ablation_trigger_signal}.

\textbf{Ranking and aggregation are complementary; which one wins alone is model-dependent.} On Qwen3-32B a single ALC-selected trajectory (Lookahead-Rank, 60.0\%) beats a consensus over unranked candidates (Adapt-Vote, 56.6\%), so aggregation alone does not drive the gain. On DeepSeek-R1-8B neither suffices alone---64.0\% for voting, 61.9\% for ranking---against 69.3\% for the full framework: lookahead prunes low-ALC branches, voting resolves what survives.

\textbf{Triggering alone is not enough.} Greedy vs.\ Blind-Sample gives inconsistent gains (+4.9 on Qwen, +0.1 on DeepSeek): locating uncertainty does not by itself separate viable from deceptive continuations.

\subsection{Scaling analysis} \label{sec:scaling_analysis}

On AIME25 with DeepSeek-R1-8B (Figure~\ref{fig:scaling_sub}), accuracy is non-decreasing in the lookahead horizon $H$, which controls how much future context scores a branch: extending from $H{=}1$ to $H{=}16$ yields a \textbf{+10.0-point} gain (73.3\% to 83.3\%) at nearly unchanged token cost (right axis). The sweep also partly isolates the ranking signal: at $H{=}1$ ALC reduces to local-confidence ranking, so the $H$ axis contrasts myopic against future-aware ranking at a near-matched budget.

\subsection{When internal confidence stops helping}
\label{sec:gpqa_boundary}

Both the trigger and ALC are internal-consistency signals, raising the question of knowledge-intensive tasks, where a trace can be \emph{confidently} wrong. On GPQA-Diamond the mathematics advantage does not transfer: \texttt{DeepConf-low} reaches 71.7\% with DeepSeek-R1-8B and 73.0\% with Qwen3-32B, while \method{} reaches 64.6\% and 72.7\%---at a much lower token budget but with no accuracy gain and a regression on the smaller model (Appendix Table~\ref{tab:gpqa_results}). A case analysis (Appendix~\ref{sec:appendix-examples}) illustrates why: a hallucinated fact yields an internally consistent continuation that receives \emph{high} ALC, which lookahead cannot veto. Internal confidence localizes instability, not falsehood.

\section{Conclusion}

\method{} allocates test-time reasoning compute at uncertainty bottlenecks: it monitors segment-level confidence, branches at detected drops, ranks candidates by ALC, and votes over a pruned subset. Across four mathematics benchmarks and four models it matches or exceeds \texttt{DeepConf-low} in 12 of 16 settings at 87.3\% fewer generated tokens. Test-time scaling can thus be made compute-aware by intervening at uncertain decisions rather than repeating whole traces---subject to the knowledge-intensive boundary observed on GPQA-Diamond.

\clearpage
\section*{Limitations}
\method{} needs white-box access to token-level log-probabilities, so it does not apply to API-only models. Our accuracy--cost comparison covers only competition mathematics with verifiable \texttt{\textbackslash boxed\{\}} answers, with 30 problems per benchmark; at that scale a sign test does not separate \method{} from \texttt{DeepConf-low} on accuracy ($p{=}0.118$), so what we defend is parity at an 87.3\% lower token budget. Token counts also exclude latency and KV-cache memory, both of which grow with the number of parallel branches. Finally, ALC scores distributional stability rather than correctness, and can prefer confidently stated but factually wrong reasoning (\S\ref{sec:gpqa_boundary}).

\section*{Acknowledgments}
Zefeng Wang is supported by the Helmholtz Association under the joint research school
``Munich School for Data Science -- MUDS''.
The authors gratefully acknowledge the computing time provided on the high-performance
computer HoreKa by the National High-Performance Computing Center at KIT (NHR@KIT).
This center is jointly supported by the Federal Ministry of Research, Technology and Space
and the Ministry of Science, Research and the Arts of Baden-Württemberg, as part of the
National High-Performance Computing (NHR) joint funding program.
HoreKa is partly funded by the German Research Foundation (DFG).

\bibliography{references}

\clearpage
\restorepaperfloatspacing
\makeatletter
\setlength{\@fptop}{0pt}
\setlength{\@fpsep}{12pt plus 4pt}
\setlength{\@fpbot}{0pt plus 1fil}
\makeatother
\appendix
\section{Algorithm and experimental setup}
\label{sec:appendix-method-algorithm}

This section provides the complete \method{} algorithm (Algorithm~\ref{alg:deeplook_overview_min}), the default hyperparameter configuration (Table~\ref{tab:deeplook_hparams}), and the prompt templates (Figures~\ref{fig:prompt-seg-pure}--\ref{fig:prompt-seg-pure-gpt}) used in experiments.

\subsection{\method{} algorithm}

Algorithm~\ref{alg:deeplook_overview_min} gives a self-contained pseudocode summary of \method{}, formalizing the five-stage pipeline described in \S\ref{sec:method}: fixed-length segment decoding, segment-level confidence scoring, local adaptive uncertainty triggering, ALC-ranked branching with adaptive pruning, and majority-vote answer synthesis.

\paragraph{Notation.}
$x$ is the input prompt and $P$ denotes the mutable decoding prefix/trace initialized from $x$.
$L_{\mathrm{main}}$ is the segment length used for main-path confidence monitoring, $L_{\mathrm{roll}}$ is the length of the rollout segment each candidate branch first generates, $L_{\mathrm{look}}$ is the shorter segment length used inside lookahead rollouts, $K$ is the branch width (number of candidate continuations sampled at each intervention), $R_{\max}$ is the maximum number of branching interventions permitted along the main decoding path, and $B = K \times R_{\max}$ is the \emph{total branch budget}.
These quantities determine the compute cost of the branching stage; all other hyperparameters govern the uncertainty trigger and are listed in Table~\ref{tab:deeplook_hparams}.

\begin{algorithm}[!t]
\caption{\method{}}
\label{alg:deeplook_overview_min}
\small
\begin{algorithmic}[1]
\State \textbf{Input:} prompt $x$; lengths $L_{\mathrm{main}}, L_{\mathrm{roll}}, L_{\mathrm{look}}$; width $K$; rounds $R_{\max}$; horizon $H$; margin $\delta$; token budget $T_{\max}$
\State \textbf{Output:} final answer $\hat{a}$
\State $P \gets x,\; \mathcal{T} \gets \emptyset,\; \mathcal{H} \gets \emptyset,\; r \gets 0$
\While{no EOS or boxed answer, and $<T_{\max}$ tokens generated}
    \algphase{Segment Monitoring:}
    \State $S \gets$ decode a candidate main-path segment from $P$ with length $L_{\mathrm{main}}$
    \State $\bar{C}(S) \gets$ segment confidence of $S$; $\tau \gets \mathrm{Percentile}(\mathcal{H}, q)$
    \Statex \hfill  $\triangleright$ Eqs.~\eqref{eq:token_conf_min}--\eqref{eq:adaptive_threshold}
    \algphase{Uncertainty Trigger:}
    \If{uncertain per Eq.~\eqref{eq:explore_trigger} \textbf{and} $r < R_{\max}$} \Comment{hysteresis state; budget remaining}
        \State $r \gets r + 1$
        \algphase{Lookahead Evaluation:}
        \State $\mathcal{B} \gets$ sample $K$ rollout segments of length $L_{\mathrm{roll}}$ from prefix $P$
        \ForAll{$b \in \mathcal{B}$}
            \State Extend $b$ by $H$ lookahead segments of length $L_{\mathrm{look}}$ (scoring only)
            \State Compute $\mathrm{ALC}(b)$ and rollout confidence $c_{\mathrm{rollout}}(b)$
        \EndFor
        \State Rank $\mathcal{B}$ in descending order of $\mathrm{ALC}$ \Comment{Eq.~\eqref{eq:alc}}
        \algphase{Primary Branch Selection:}
        \State $\mathcal{Q} \gets \{b \in \mathcal{B}: \mathrm{ALC}(b) \geq \tau \wedge \mathrm{ALC}(b) > \bar{C}(S)\}$
        \If{$\mathcal{Q} \neq \emptyset$}
            \State $b^* \gets \arg\max_{b \in \mathcal{Q}} \mathrm{ALC}(b)$
        \Else
            \State $b^* \gets \arg\max_{b \in \mathcal{B}} c_{\mathrm{rollout}}(b)$
        \EndIf
        \algphase{Adaptive Pruning:}
        \State Compute confidence gap $\Delta_i \gets \max(0, \tau - \bar{C}(S))$ and keep ratio $\rho_i$
        \Statex \hfill  $\triangleright$ Eq.~\eqref{eq:dyn_keep_simple}
        \State $\mathcal{B}_{\mathrm{keep}} \gets$ top-$\lfloor K\rho_i \rfloor$ ALC-ranked branches of $\mathcal{B}$, minus $b^*$
        \State Complete each $b \in \mathcal{B}_{\mathrm{keep}}$ with main-path sampling; add to $\mathcal{T}$
        \State $P \gets P \oplus b^*$ \Comment{$b^*$'s rollout replaces $S$}
    \Else
        \State $P \gets P \oplus S$ \Comment{$\mathcal{H}$ receives $\bar{C}(S)$ in both cases}
    \EndIf
\EndWhile
\algphase{Answer Synthesis:}
\State $\mathcal{T} \gets \mathcal{T} \cup \{P\}$
\State $\hat{a} \gets \arg\max_a \sum_{t \in \mathcal{T}} \mathbb{I}(\text{Answer}(t)=a)$ \Comment{Eq.~\eqref{eq:majority_vote}}
\State \Return $\hat{a}$
\end{algorithmic}
\end{algorithm}

\subsection{Generation hyperparameters}
\label{sec:appendix-gen-hparams}

Table~\ref{tab:deeplook_hparams} lists the default hyperparameters shared across all four model families.
The three experimental variants differ only in their total branch budget $B = K \times R_{\max}$ (see Algorithm~\ref{alg:deeplook_overview_min}): with $R_{\max}=2$ fixed, \texttt{\method{}@32}, \texttt{\method{}@64}, and \texttt{\method{}@128} set $K$ to 16, 32, and 64, giving $B=32$, $64$, and $128$, respectively.
Sensitivity to the trigger parameters ($q$, $W$, $\delta$, $k$) is analyzed in Appendix~\ref{sec:ablation_hyper}, and sensitivity to $R_{\max}$ and the keep ratio in Appendix~\ref{sec:ablation_hyperparams}.

\begin{table}[t]
  \centering
  \small
  \setlength{\tabcolsep}{6pt}
  \renewcommand{\arraystretch}{1.05}
  \begin{tabular}{>{\raggedright\arraybackslash}p{0.55\columnwidth}
                  >{\raggedright\arraybackslash}p{0.36\columnwidth}}
    \toprule
    Hyperparameter & Value \\
    \midrule
    Main segment length $L_{\mathrm{main}}$ & 512 \\
    Confidence top-$k$ & 20 \\
    Trigger window $W$ & 8 segments \\
    Low quantile $q$ & 0.10 \\
    Warmup & 4 segments \\
    Hysteresis $\delta$ & 0.02 \\
    Branch width $K$ & 16 \\
    Rollout segment length $L_{\mathrm{roll}}$ & 512 \\
    Lookahead horizon $H$ & 16 \\
    Lookahead segment length $L_{\mathrm{look}}$ & 32 \\
    Keep ratio $\rho_i$ & dynamic in $[0.10, 0.25]$ \\
    Base keep ratio $\rho_{\text{base}}$ & 0.10 \\
    Keep-gap scale $s$ & 0.03 \\
    Max branch rounds $R_{\max}$ & 2 \\
    \bottomrule
  \end{tabular}
  \caption{\textbf{Default \method{} hyperparameters} (corresponding to \texttt{\method{}@32}).
  The \texttt{@64} and \texttt{@128} variants increase only $K$ (to 32 and 64); all other values are unchanged.}
  \label{tab:deeplook_hparams}
\end{table}

\subsection{Prompt templates}
\label{sec:appendix-prompts}

We use two prompt templates depending on the model family. DeepSeek and Qwen models receive a plain chat-format instruction (Figure~\ref{fig:prompt-seg-pure}), while GPT-OSS models additionally configure \texttt{reasoning\_effort} via the tokenizer template (Figure~\ref{fig:prompt-seg-pure-gpt}). Both templates enforce chain-of-thought reasoning with a final \texttt{\textbackslash boxed\{\}} answer, which the voting logic in \S\ref{sec:adaptive_pruning} depends on.

\begin{figure}[t]
\centering
\begin{tcolorbox}[
  enhanced,
  width=\linewidth,
  colback=white,
  colframe=black!25,
  boxrule=0.4pt,
  arc=1.5mm,
  left=2mm, right=2mm, top=3mm, bottom=2mm,
  title=\textbf{\small Prompt A.1: Pure Segment Decoding (DeepSeek/Qwen)},
  coltitle=white,
  colbacktitle=black!60,
  attach boxed title to top left={yshift=-2mm, xshift=2mm},
  boxed title style={arc=1mm, boxrule=0pt},
]
\small
\textit{Purpose.} This prompt is prepended to each math query to enforce step-by-step reasoning and a final \texttt{\textbackslash boxed\{\}} answer format.

\vspace{1mm}
{\ttfamily\small\raggedright
\textbf{[System]} \\
(DeepSeek only) This assistant is DeepSeek-R1, created by DeepSeek. \\[1mm]
\textbf{[User]} \\
\{question\} \\[1mm]
Please reason step by step, and put your final answer within \texttt{\textbackslash boxed\{\}}.
}
\end{tcolorbox}
\caption{Prompt logic for \texttt{model\_type=deepseek/qwen}.}
\label{fig:prompt-seg-pure}
\end{figure}

\begin{figure}[t]
\centering
\begin{tcolorbox}[
  enhanced,
  width=\linewidth,
  colback=white,
  colframe=black!25,
  boxrule=0.4pt,
  arc=1.5mm,
  left=2mm, right=2mm, top=3mm, bottom=2mm,
  title=\textbf{\small Prompt A.2: Pure Segment Decoding (GPT Template)},
  coltitle=white,
  colbacktitle=black!60,
  attach boxed title to top left={yshift=-2mm, xshift=2mm},
  boxed title style={arc=1mm, boxrule=0pt},
]
\small
\textit{Purpose.} Same instruction as Figure~\ref{fig:prompt-seg-pure}, with optional \texttt{reasoning\_effort}.

\vspace{1mm}
{\ttfamily\small\raggedright
\textbf{[User]} \\
\{question\} \\[1mm]
Please reason step by step, and put your final answer within \texttt{\textbackslash boxed\{\}}. \\[2mm]
\textbf{[Template Configuration]} \\
tokenizer.apply\_chat\_template( \\
\hspace*{1em}messages, tokenize=False, \\
\hspace*{1em}reasoning\_effort=\{high|medium|low\}, \\
\hspace*{1em}add\_generation\_prompt=True)
}
\end{tcolorbox}
\caption{Prompt logic for GPT-OSS models with \texttt{reasoning\_effort}.}
\label{fig:prompt-seg-pure-gpt}
\end{figure}

\subsection{Existing asset licenses}
\label{sec:asset_licenses}

Table~\ref{tab:asset_licenses} lists the third-party datasets and model checkpoints used in our experiments, together with the license metadata reported by the corresponding public asset pages at the time of access.
We use these assets only for inference-time evaluation and do not redistribute, modify, or repackage the datasets or model weights.
Baseline methods are credited through the cited papers in \S\ref{sec:baselines}; we implement the inference procedures ourselves rather than incorporating external baseline code.

\begin{table}[t]
  \centering
  \scriptsize
  \setlength{\tabcolsep}{3.5pt}
  \renewcommand{\arraystretch}{1.1}
  \resizebox{\columnwidth}{!}{
  \begin{tabular}{p{0.16\columnwidth}p{0.29\columnwidth}p{0.27\columnwidth}p{0.20\columnwidth}}
    \toprule
    Asset type & Asset & Source/version used & License or terms \\
    \midrule
    Dataset & AIME24~\citep{jia_aime24_2024} & \url{https://huggingface.co/datasets/Maxwell-Jia/AIME_2024}, 2024 dataset card & MIT License \\
    Dataset & AIME25~\citep{matharena_aime25_2025} & \url{https://huggingface.co/datasets/MathArena/aime_2025}, 2025 MathArena dataset card & CC BY-NC-SA 4.0 \\
    Dataset & BRUMO25~\citep{matharena_brumo25_2025} & \url{https://huggingface.co/datasets/MathArena/brumo_2025}, 2025 MathArena dataset card & CC BY-NC-SA 4.0 \\
    Dataset & HMMT25~\citep{matharena_hmmt25_2025} & \url{https://huggingface.co/datasets/MathArena/hmmt_feb_2025}, 2025 MathArena dataset card & CC BY-NC-SA 4.0 \\
    Model & DeepSeek-R1-8B~\citep{deepseek_r1} & \url{https://huggingface.co/deepseek-ai/DeepSeek-R1-0528-Qwen3-8B} & MIT License \\
    Model & Qwen3-32B~\citep{qwen3_techreport} & \url{https://huggingface.co/Qwen/Qwen3-32B} & Apache License 2.0 \\
    Model & GPT-OSS-20B~\citep{openai_gptoss} & \url{https://huggingface.co/openai/gpt-oss-20b} & Apache License 2.0 \\
    Model & GPT-OSS-120B~\citep{openai_gptoss} & \url{https://huggingface.co/openai/gpt-oss-120b} & Apache License 2.0 \\
    \bottomrule
  \end{tabular}}
  \caption{\textbf{Licenses and terms for existing assets.}
  Dataset and model licenses are taken from the linked public asset pages.}
  \label{tab:asset_licenses}
\end{table}

\section{Additional experimental results}
\label{sec:main_results_ap}

\begin{figure*}[t]
    \centering
    \includegraphics[width=\textwidth]{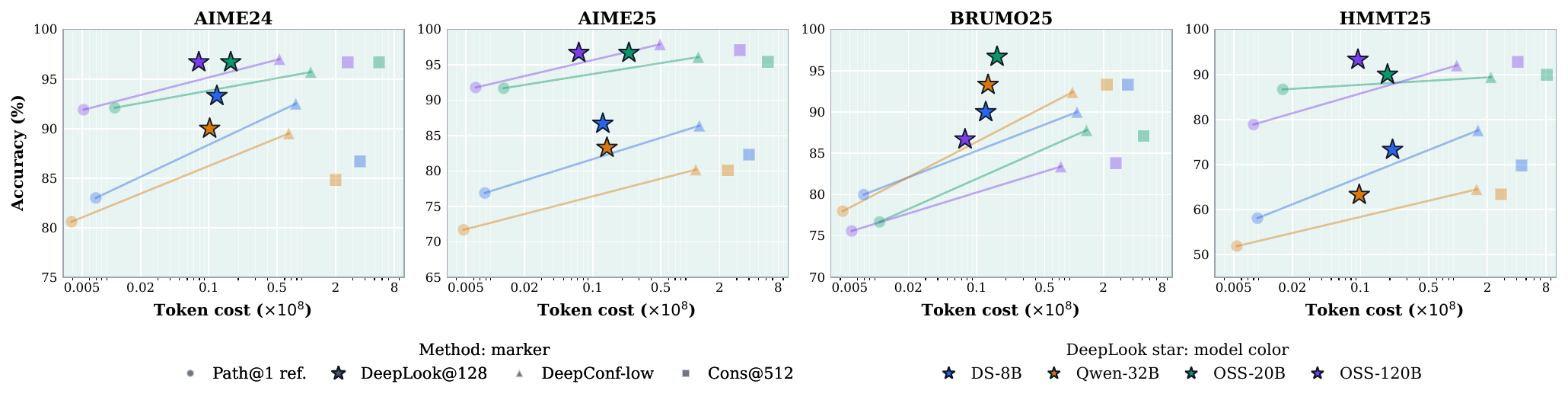}
    \caption{\textbf{Accuracy vs. token cost trade-offs across benchmarks.}
    Each panel plots dataset-level token cost against accuracy for one benchmark.
    Colors identify the base model, with \method{}@128 shown as prominent stars and the corresponding \texttt{Path@1}, \texttt{DeepConf-low}, and \texttt{Cons@512} references shown with lighter markers.
    Faint colored segments connect \texttt{Path@1} to \texttt{DeepConf-low} within the same model, visualizing the baseline accuracy--cost frontier that \method{}@128 pushes beyond.
    Across AIME24, AIME25, BRUMO25, and HMMT25, \method{}@128 consistently lies in the low-cost, high-accuracy region: it uses far fewer tokens than \texttt{DeepConf-low} and \texttt{Cons@512}, while matching or exceeding their accuracy in most model--dataset settings.
    \texttt{Path@1} is shown as the no-search reference point, plotted at the mean per-trajectory cost implied by \texttt{Cons@512} (its token total divided by 512), which is the cost matching a single sampled trajectory.}
    \label{fig:main_results_pareto_by_dataset}
\end{figure*}

We report complete accuracy and token-cost results across four benchmarks and four model families \citep{deepseek_r1, qwen3_techreport, openai_gptoss}. Table~\ref{tab:main_results_full} extends the main-text comparison by adding the \texttt{DeepConf-high} configuration and three \method{} budgets. The \texttt{\method{}@32}, \texttt{\method{}@64}, and \texttt{\method{}@128} variants differ only in branch width $K$; all other hyperparameters are held fixed at the values in Table~\ref{tab:deeplook_hparams}.
Table~\ref{tab:gpqa_results} reports the GPQA-Diamond comparison discussed in \S\ref{sec:gpqa_boundary}.

\begin{table*}[t]
  \centering
  \scriptsize
  \setlength{\tabcolsep}{3.2pt}
  \renewcommand{\arraystretch}{1.02}
  \resizebox{\textwidth}{!}{
  \begin{tabular}{llcccccc|cccccc}
    \toprule
    \multirow{2}{*}{Model} & \multirow{2}{*}{Dataset} &
    \multicolumn{2}{c}{Cons@512} &
    \multicolumn{2}{c}{DeepConf-high} &
    \multicolumn{2}{c|}{DeepConf-low} &
    \multicolumn{2}{c}{\method{}@32} &
    \multicolumn{2}{c}{\method{}@64} &
    \multicolumn{2}{c}{\method{}@128} \\
    \cmidrule(lr){3-4}\cmidrule(lr){5-6}\cmidrule(lr){7-8}
    \cmidrule(lr){9-10}\cmidrule(lr){11-12}\cmidrule(lr){13-14}
    & & Tok & Acc & Tok & Acc & Tok & Acc & Tok & Acc & Tok & Acc & Tok & Acc \\
    \midrule
    \multirow{4}{*}{DeepSeek-R1-8B}
      & AIME24  & 3.55 & 86.7\% & 1.45 & 86.7\% & 0.78 & 92.5\% & 0.036 & 86.7\% & 0.064 & 86.7\% & 0.121 & 93.3\% \\
      & AIME25  & 4.01 & 82.3\% & 2.37 & 81.4\% & 1.24 & 86.4\% & 0.027 & 83.3\% & 0.072 & 83.3\% & 0.127 & 86.7\% \\
      & BRUMO25 & 3.56 & 93.3\% & 2.17 & 93.3\% & 1.07 & 90.0\% & 0.033 & 86.7\% & 0.065 & 90.0\% & 0.124 & 90.0\% \\
      & HMMT25  & 4.49 & 69.8\% & 3.43 & 70.0\% & 1.60 & 77.6\% & 0.040 & 66.7\% & 0.082 & 70.0\% & 0.213 & 73.3\% \\
    \midrule
    \multirow{4}{*}{Qwen3-32B}
      & AIME24  & 2.00 & 84.8\% & 0.88 & 86.4\% & 0.66 & 89.5\% & 0.018 & 86.7\% & 0.046 & 86.7\% & 0.102 & 90.0\% \\
      & AIME25  & 2.43 & 80.1\% & 1.61 & 80.2\% & 1.14 & 80.2\% & 0.038 & 80.0\% & 0.068 & 80.0\% & 0.140 & 83.3\% \\
      & BRUMO25 & 2.17 & 93.3\% & 1.37 & 93.3\% & 0.96 & 92.4\% & 0.021 & 86.7\% & 0.063 & 90.0\% & 0.131 & 93.3\% \\
      & HMMT25  & 2.76 & 63.4\% & 2.24 & 63.6\% & 1.55 & 64.5\% & 0.028 & 56.7\% & 0.047 & 60.0\% & 0.097 & 63.3\% \\
    \midrule
    \multirow{4}{*}{GPT-OSS-20B}
      & AIME24  & 5.57 & 96.7\% & 3.07 & 96.7\% & 1.11 & 95.7\% & 0.044 & 93.3\% & 0.086 & 93.3\% & 0.168 & 96.7\% \\
      & AIME25  & 6.26 & 95.4\% & 3.18 & 95.3\% & 1.21 & 96.1\% & 0.054 & 86.7\% & 0.102 & 93.3\% & 0.235 & 96.7\% \\
      & BRUMO25 & 5.16 & 87.1\% & 3.49 & 87.2\% & 1.34 & 87.8\% & 0.042 & 93.3\% & 0.095 & 96.7\% & 0.162 & 96.7\% \\
      & HMMT25  & 8.16 & 89.9\% & 6.03 & 90.3\% & 2.17 & 89.4\% & 0.072 & 90.0\% & 0.126 & 90.0\% & 0.189 & 90.0\% \\
    \midrule
    \multirow{4}{*}{GPT-OSS-120B}
      & AIME24  & 2.66 & 96.7\% & 1.20 & 96.7\% & 0.53 & 97.0\% & 0.022 & 93.3\% & 0.042 & 96.7\% & 0.079 & 96.7\% \\
      & AIME25  & 3.23 & 97.1\% & 1.42 & 97.0\% & 0.49 & 97.9\% & 0.034 & 90.0\% & 0.045 & 96.7\% & 0.072 & 96.7\% \\
      & BRUMO25 & 2.68 & 83.8\% & 1.81 & 84.0\% & 0.73 & 83.4\% & 0.019 & 83.3\% & 0.049 & 86.7\% & 0.076 & 86.7\% \\
      & HMMT25  & 4.09 & 92.8\% & 2.78 & 93.0\% & 0.97 & 92.0\% & 0.036 & 86.7\% & 0.072 & 86.7\% & 0.094 & 93.3\% \\
    \bottomrule
  \end{tabular}}
  \caption{\textbf{Full benchmark results.}
  Accuracy (\%) and token cost (Tok, $\times 10^8$) for \textbf{Cons@512}, DeepConf baselines, and \method{} budgets.}
  \label{tab:main_results_full}
\end{table*}

\begin{table}[t]
  \centering
  \scriptsize
  \setlength{\tabcolsep}{2.5pt}
  \begin{tabular}{lccccccc}
    \toprule
    & Path@1 & \multicolumn{2}{c}{Cons@512} & \multicolumn{2}{c}{DeepConf-low} & \multicolumn{2}{c}{\method{}@128} \\
    \cmidrule(lr){3-4}\cmidrule(lr){5-6}\cmidrule(lr){7-8}
    Model & Acc & Tok & Acc & Tok & Acc & Tok & Acc \\
    \midrule
    DeepSeek-R1-8B & 62.8\% & 10.9 & 66.2\% & 3.46 & 71.7\% & 0.198 & 64.6\% \\
    Qwen3-32B      & 68.9\% & 7.44 & 72.2\% & 3.21 & 73.0\% & 0.376 & 72.7\% \\
    \bottomrule
  \end{tabular}
  \caption{\textbf{GPQA-Diamond results} (\S\ref{sec:gpqa_boundary}).
  Accuracy (\%) and token cost (Tok, $\times 10^8$).}
  \label{tab:gpqa_results}
\end{table}

\section{Token cost breakdown}
\label{sec:token_breakdown}

Table~\ref{tab:token_breakdown} decomposes \method{}'s generated-token budget on AIME25 with DeepSeek-R1-8B into four inference stages: main-path decoding, branch rollouts, ALC lookahead evaluation, and completion of retained branches. ALC scoring accounts for 23.58\% of generated tokens, while retained-branch completion is the largest component at 39.52\%. The lookahead mechanism is therefore not the dominant cost; most computation is spent after pruning, on branches that ALC has judged worth completing.

This breakdown also clarifies the scope of our efficiency metric: token count captures generated-token work, not wall-clock latency. In a parallel implementation, evaluating $K$ branches simultaneously can reduce elapsed time, but increases KV-cache memory proportionally with the number of active branches---leaving latency--memory trade-offs as a deployment-level concern.

\begin{table}[t]
  \centering
  \small
  \setlength{\tabcolsep}{5pt}
  \begin{tabular}{lr}
    \toprule
    Component & \% of Total \\
    \midrule
    Main-path decoding      & 13.25\% \\
    Branch rollouts         & 23.65\% \\
    Lookahead evaluation    & 23.58\% \\
    Retained-branch completions & 39.52\% \\
    \midrule
    \textbf{Total}          & 100.00\% \\
    \bottomrule
  \end{tabular}
  \caption{\textbf{Token cost breakdown on AIME25 (DeepSeek-R1-8B).}
  Stage-wise share of \method{}'s dataset-level generated-token budget.
  The lookahead evaluation (ALC scoring) accounts for under 24\% of total tokens.}
  \label{tab:token_breakdown}
\end{table}

\section{Additional uncertainty analysis}
\label{sec:appendix-qwen-kde}

\begin{figure}[t]
  \centering
  \includegraphics[width=\columnwidth, trim={0 9cm 0 1cm}, clip]{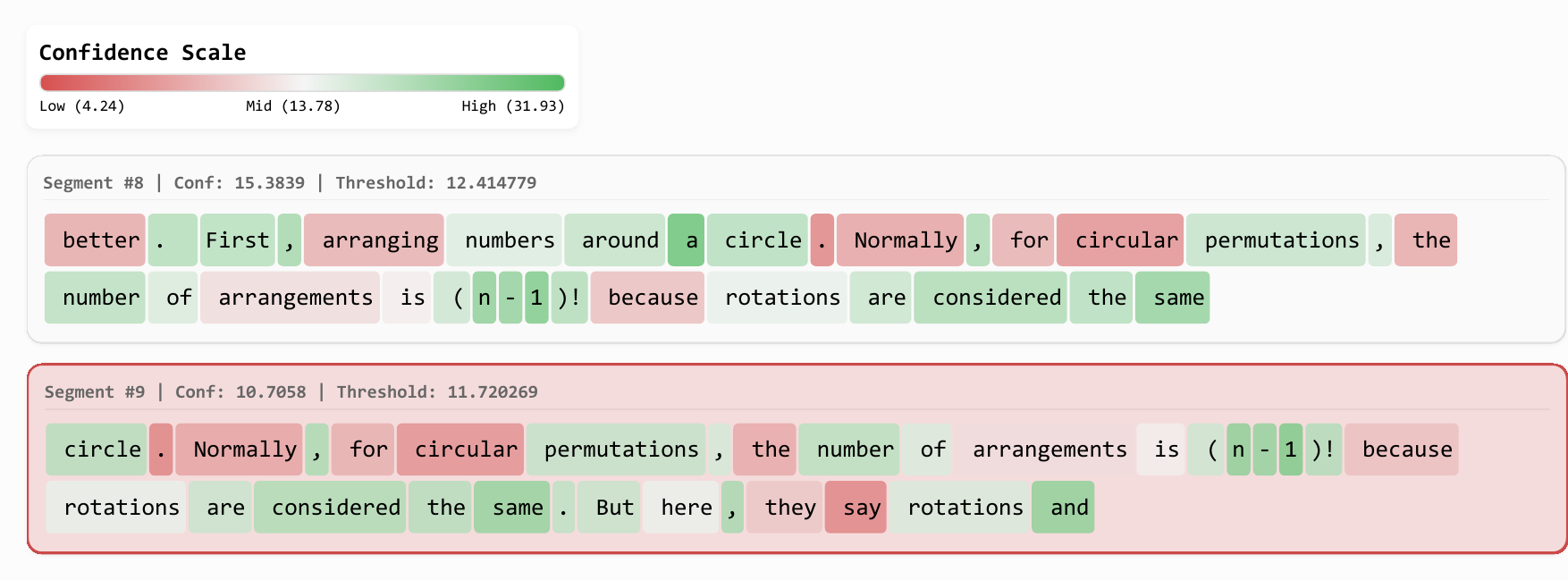}
  \caption{
  \textbf{Token- vs.\ segment-level confidence.}
  Each segment spans two lines for readability (segment length $32$ tokens for illustration).
  The segment background encodes segment-level confidence $\bar{C}$,
  where red backgrounds indicate uncertain segments.
  Individual token colors ($C_t$) reveal local confidence fluctuations,
  motivating segment-level aggregation for robust detection of reasoning failures.
  }
  \label{fig:combined_conf_vis}
\end{figure}

\begin{figure}[t]
  \centering
  \begin{subfigure}{0.49\linewidth}
    \centering
    \includegraphics[width=\linewidth]{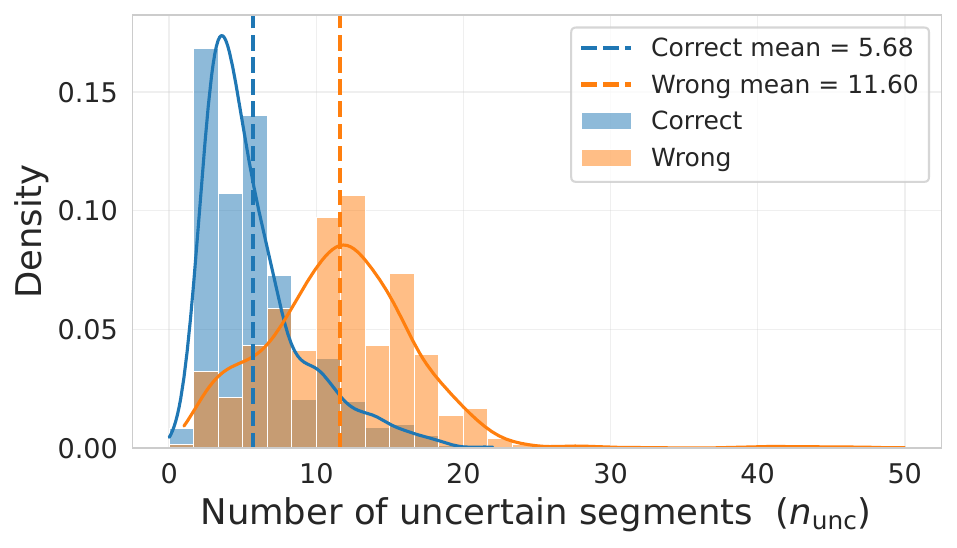}
    \caption{$n_{\text{unc}}$ distribution.}
    \label{fig:appendix-kde-qwen-nunc}
  \end{subfigure}
  \hfill
  \begin{subfigure}{0.49\linewidth}
    \centering
    \includegraphics[width=\linewidth]{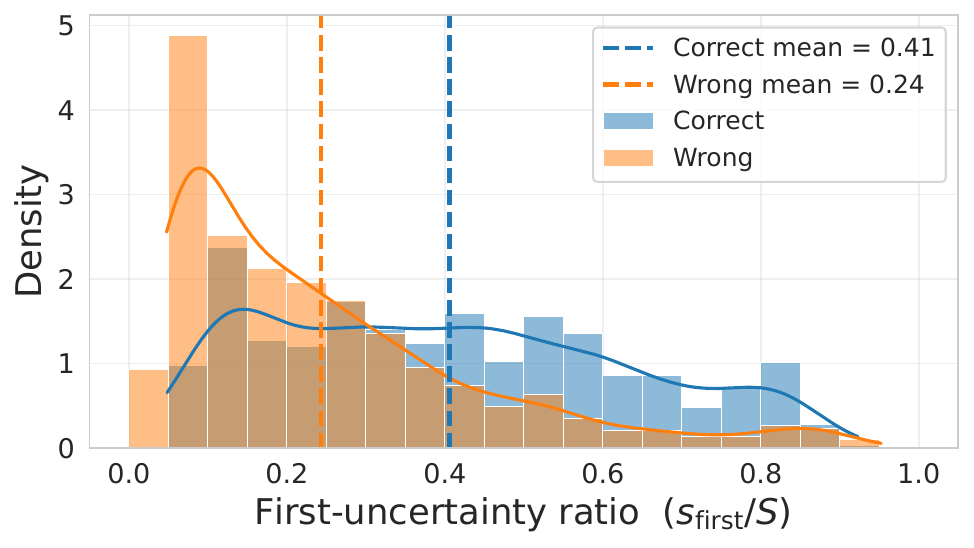}
    \caption{$t_{\text{first}}/N_{\text{seg}}$ distribution.}
    \label{fig:appendix-kde-qwen-tfirst}
  \end{subfigure}
  \caption{\textbf{Motivating analysis on Qwen3-32B.}
  Incorrect traces show both more frequent uncertainty and earlier uncertainty onset than correct traces.}
  \label{fig:appendix-kde-qwen}
\end{figure}

Figure~\ref{fig:appendix-kde-qwen} confirms that Qwen3-32B follows the same uncertainty pattern as DeepSeek-R1-8B reported in the main text: incorrect traces contain more uncertain segments on average (11.60 vs.\ 5.68) and exhibit an earlier first confidence drop (position ratio 0.24 vs.\ 0.41). The signal is therefore consistent across both model families examined, which are architecturally distinct, supporting segment-level confidence as a trigger signal.

\section{Additional ablation studies}
\label{sec:ablation}

We include three ablations that extend the component and scaling analysis in \S\ref{sec:experiments}: (i) a comparison against recursive tree expansion (\S\ref{sec:ablation_strategy}); (ii) a one-at-a-time sensitivity study over trigger hyperparameters (\S\ref{sec:ablation_hyper}), showing that the threshold parameters are robust while segment granularity is not; and (iii) a sensitivity study over branching rounds and keep ratio (\S\ref{sec:ablation_hyperparams}), quantifying the accuracy--cost trade-off as the completion budget grows. The signal-level analysis of the confidence trigger appears in the main text (\S\ref{sec:ablation_trigger_signal}).

\subsection{Search strategy: \method{} vs. \texttt{DeepLook-Recursive}}
\label{sec:ablation_strategy}

\method{} uses asymmetric exploration: only the main path remains under active confidence monitoring, while side branches are treated as non-recursive lookahead rollouts. We compare this design with \texttt{DeepLook-Recursive}, a recursive variant of \method{} that monitors and expands every generated branch.

Table~\ref{tab:deeplook_recursive_vs_deeplook32} shows that \method{} achieves a better accuracy--token-cost trade-off across both Qwen3-32B benchmarks: at essentially matched token budgets, \method{} is more accurate on both. Expanding uncertainty-driven interventions onto already-speculative side paths appears to add noise rather than correct it; \method{}'s asymmetric design concentrates active monitoring on the primary trajectory and treats side branches as fixed-horizon rollouts.

\begin{table}[t]
  \centering
  \small
  \setlength{\tabcolsep}{3pt}
  \begin{tabular}{lcccc}
    \toprule
    & \multicolumn{2}{c}{\texttt{DeepLook-Recursive}} & \multicolumn{2}{c}{\method{}} \\
    \cmidrule(lr){2-3}\cmidrule(lr){4-5}
    Dataset & Tok & Acc & Tok & Acc \\
    \midrule
    AIME24 & 0.019 & 83.3 & \textbf{0.018} & \textbf{86.7} \\
    AIME25 & 0.039 & 76.7 & \textbf{0.038} & \textbf{80.0} \\
    \bottomrule
  \end{tabular}
  \caption{\textbf{Search strategy: \texttt{DeepLook-Recursive} vs.\ \method{}.}
  Tok ($\times 10^8$) and Acc (\%) on Qwen3-32B.}
  \label{tab:deeplook_recursive_vs_deeplook32}
\end{table}

\subsection{Trigger parameter sensitivity}
\label{sec:ablation_hyper}

A key concern for an uncertainty-driven method is whether performance depends critically on how the trigger is tuned.
Table~\ref{tab:hyper_sensitivity} reports a one-at-a-time sensitivity analysis on AIME25 (DeepSeek-R1-8B), varying one parameter at a time around a fixed ablation baseline.

\begin{table}[t]
  \centering
  \small
  \setlength{\tabcolsep}{3.5pt}
  \renewcommand{\arraystretch}{1.0}
  \begin{tabular}{llccc}
    \toprule
    Hyperparameter & Setting & Acc (\%) & Tok (M) & Trig. \\
    \midrule
    \multicolumn{2}{l}{\emph{Ablation baseline}} & 83.3 & 5.0 & 5.5 \\
    \midrule
    \multicolumn{5}{l}{\emph{Robust: accuracy within one problem of baseline}} \\
    \multirow{2}{*}{Low quantile $q$}
      & 0.05 & 80.0 & 4.5 & 4.2 \\
      & 0.25 & 80.0 & 7.5 & 7.3 \\
    \multirow{2}{*}{Window size $W$}
      & 4    & 80.0 & 5.5 & 6.4 \\
      & 32   & 80.0 & 6.9 & 8.5 \\
    \multirow{2}{*}{Hysteresis $\delta$}
      & 0    & 83.3 & 5.8 & 6.8 \\
      & 0.20 & 80.0 & 7.5 & 9.2 \\
    Confidence top-$k$ & 20 & 83.3 & 4.9 & 5.3 \\
    \midrule
    \multicolumn{5}{l}{\emph{Sensitive: what defines a segment}} \\
    Segment length $L_{\mathrm{main}}$ & 1024 & 76.7 & 4.6 & 3.1 \\
    Warm-up (segments) & 16 & 76.7 & 4.3 & 3.2 \\
    \bottomrule
  \end{tabular}
  \caption{\textbf{Trigger parameter sensitivity on AIME25 (DeepSeek-R1-8B).} One parameter varied at a time; all others held at the ablation baseline ($q{=}0.10$, $W{=}8$, $\delta{=}0.02$, $k{=}10$, $L_{\mathrm{main}}{=}512$, warm-up $4$, $K{=}16$, $R_{\max}{=}10$). This ablation baseline differs from the default configuration of Table~\ref{tab:deeplook_hparams} in $k$ and $R_{\max}$, so its absolute token counts are not comparable to those of Table~\ref{tab:ablation_combined_hyper}. Tok in millions; ``Trig.'' is the mean number of trigger firings per problem.}
  \label{tab:hyper_sensitivity}
\end{table}

\textbf{The four trigger parameters are robust.} Across $q$, $W$, $\delta$ and top-$k$, accuracy stays within one problem of the baseline: what these parameters buy is compute, not accuracy.
(1) \emph{$q$ is a compute--accuracy dial.} Reducing $q$ to 0.05 causes occasional missed early errors; raising it to 0.25 over-explores minor lexical hesitations, inflating token cost by 50\% with no accuracy benefit. Behavior degrades smoothly within $[0.05, 0.25]$---never catastrophically.
(2) \emph{$W$ sets how much history the threshold remembers.} A large window ($W{=}32$) keeps high-confidence scores from early, easy segments in the reference set, which raises the low quantile and makes the trigger fire \emph{more} often later (8.5 vs.\ 5.5 firings per problem) at 38\% higher cost, without improving accuracy.
(3) \emph{$\delta$ debounces recovery.} With $\delta{=}0$ the uncertain state releases as soon as confidence returns to the threshold, and minor fluctuations immediately re-trigger (6.8 firings per problem versus 5.5); a large margin ($\delta{=}0.20$) instead holds the trigger latched far longer (9.2 firings). $\delta{=}0.02$ suppresses both failure modes without dulling sensitivity to genuine drops.
(4) \emph{top-$k$ barely matters.} $k \in \{10, 20\}$ gives near-identical accuracy and firing rate, so the confidence estimate is insensitive to candidate count.

\textbf{Segmentation itself is not robust.} The two parameters that define what a ``segment'' \emph{is} behave differently: quadrupling the main segment length ($L_{\mathrm{main}}{=}1024$) or the warm-up ($16$ segments) each costs two problems ($76.7\%$). Both act by suppressing the trigger---firings drop to $3.1$ and $3.2$ per problem---so the method loses its intervention points rather than mis-ranking branches. The practical reading is that \method{} tolerates a wide range of \emph{threshold} settings but needs the segment granularity fine enough for a confidence drop to be localized.

\subsection{Branching-round and keep-ratio sensitivity}
\label{sec:ablation_hyperparams}

Table~\ref{tab:ablation_combined_hyper} varies the maximum number of branching rounds $R_{\max}$ on AIME25 with DeepSeek-R1-8B ($K=16$). Across $R_{\max} \in \{2,4,6,8\}$, accuracy is flat at 83.3\% while token cost grows by roughly $2.4\times$: within this range the extra branches do not change the voted answer. We therefore use $R_{\max}=2$ by default---the cheapest point in this range---matching the low-budget setting reported in the main text.

Table~\ref{tab:ablation_prune_qwen} varies the static keep ratio $\rho$ on AIME25 with Qwen3-32B ($K=16$, $R_{\max}=2$). Accuracy improves once $\rho \geq 0.50$, where enough branches survive to benefit voting; below this threshold, over-aggressive pruning discards potentially correct candidates. The operative quantity is the retained-branch count rather than the ratio: at $K{=}16$, $\rho{=}0.50$ retains 8 branches, comparable to the 6--16 branches that the dynamic keep ratio of the main configuration retains at $K{=}64$. The dynamic rule adapts $\rho$ to the observed confidence gap $\Delta_i$ (Eq.~\ref{eq:dyn_keep_simple}), providing high retention only when uncertainty is large and limiting cost elsewhere.

\begin{table}[t]
  \centering
  \small
  \setlength{\tabcolsep}{5pt}
  \begin{tabular}{ccc}
    \toprule
    Rounds & Tok ($\times 10^8$) & Acc (\%) \\
    \midrule
    \textbf{2}  & 0.0270 & 83.3 \\
     4  & 0.0437 & 83.3 \\
     6  & 0.0545 & 83.3 \\
     8  & 0.0649 & 83.3 \\
    \bottomrule
  \end{tabular}
  \caption{\textbf{Branching-round sensitivity on AIME25 (DeepSeek-R1-8B).}
  $K=16$. Bold marks the default $R_{\max}$.}
  \label{tab:ablation_combined_hyper}
\end{table}

\begin{table}[t]
  \centering
  \small
  \setlength{\tabcolsep}{5pt}
  \begin{tabular}{ccc}
    \toprule
    $\rho$ & Tok ($\times 10^8$) & Acc (\%) \\
    \midrule
    0.10 & 0.0180 & 80.0 \\
    0.25 & 0.0438 & 80.0 \\
    \textbf{0.50} & 0.0850 & \textbf{83.3} \\
    0.75 & 0.1204 & \textbf{83.3} \\
    \bottomrule
  \end{tabular}
  \caption{\textbf{Keep-ratio sensitivity on AIME25 (Qwen3-32B).}
  $K=16$, $R_{\max}=2$; static keep ratio $\rho$.}
  \label{tab:ablation_prune_qwen}
\end{table}

\section{Examples (illustrative summaries)}
\label{sec:appendix-examples}

The examples below are \emph{illustrative summaries}, not verbatim model traces: complete traces are far too long for readable presentation, so we hand-condense the decisive portions of each answer---the key reasoning step where the baseline falls into an error mode and the corresponding \method{} correction that changes the final answer. Each figure is labeled accordingly; these summaries convey how the intervention plays out in the logged runs but should not be read as reproduced model output.

\begin{figure}[!t]
    \centering
    \begin{tcolorbox}[colback=gray!5!white, colframe=gray!75!black, title=\textbf{The Heuristic Trap in Combinatorial Optimization} --- \textit{Illustrative Summary}]
        \small
        \textbf{Problem (HMMT 2025, Combinatorics):} 
        In an $11 \times 11$ grid, maximize the number of doors Karthik can traverse in a closed loop without repeating edges.
        
        \textit{\textbf{Mathematical Context:} Total Edges = 220. The grid contains 36 vertices with odd degrees. A closed loop (Eulerian circuit) requires all vertices to have even degrees, necessitating edge removal to pair up odd vertices.}
        
        \rule{\textwidth}{0.4pt}
        
        \begin{minipage}[t]{0.48\textwidth}
            \textbf{\textcolor{red}{Cons@512 (Baseline Failure Path)}} \\
            \textit{Model applies a standard heuristic without verifying global constraints.}
            
            \vspace{0.1cm}
            \textbf{Step 1:} 
            Identifies 36 vertices with odd degrees. Applies the standard formula: $\text{Edges to Remove} = \frac{36}{2} = 18$. \\
            \textbf{Step 2:} 
            ``Since the grid is symmetric, local pairings are always possible regardless of boundary parity.'' \\
            \textbf{Step 3:} 
            $220 - 18 = 202$. \\
            \textbf{Step 4:} 
            Multiple paths converge on 202 because they all rely on the same incomplete heuristic. \textbf{Path Selected.}
            
            \vspace{0.1cm}
            \textbf{Final Answer:} \boxed{202} \quad \textcolor{red}{\textbf{(Incorrect)}}
        \end{minipage}
        \hfill
        \vline
        \hfill
        \begin{minipage}[t]{0.48\textwidth}
            \textbf{\textcolor{green!40!black}{\method{} (Ours --- Correct Path)}} \\
            \textit{Lookahead mechanism identifies topological constraints.}
            
            \vspace{0.1cm}
            \textbf{Step 1:} 
            Initially attempts to remove 18 edges based on the $N/2$ odd-vertex rule. \\
            \textbf{Step 2:} 
             Detects instability: ``Wait, each side has 9 odd vertices. 9 is odd, so they cannot all be paired locally.'' \\
            \textbf{Step 3:} 
            Pairing the `leftover' vertices across corners increases the cost by 2 additional edges. $\text{Remove} = 18 + 2 = 20$. \\
            \textbf{Step 4:} 
            $220 - 20 = 200$. The lookahead mechanism ensures the pairing strategy is globally consistent.
            
            \vspace{0.1cm}
            \textbf{Final Answer:} \boxed{200} \quad \textcolor{green!40!black}{\textbf{(Correct)}}
        \end{minipage}
    \end{tcolorbox}
    \caption{\textbf{Illustrative summary: \method{} exposes global constraints that self-consistency misses.} In this HMMT combinatorics example, \texttt{Cons@512} repeatedly selects the locally plausible odd-vertex heuristic and converges to 202. By intervening at the unstable pairing step, \method{} uses lookahead to test whether the proposed edge removals remain globally consistent, detects the boundary-parity obstruction, and redirects the solution to the correct answer 200.}
    \label{fig:hmmt_case_study_unified}
\end{figure}

\begin{figure}[!t]
    \centering
    \begin{tcolorbox}[colback=gray!5!white, colframe=gray!75!black, title=\textbf{The Overcounting Trap in Symmetrical Combinatorics} --- \textit{Illustrative Summary}]
        \small
        \textbf{Problem (AIME 2024, Combinatorics/Probability):} 
        Each vertex of a regular octagon is independently colored red or blue. The probability that the octagon can be rotated so that all blue vertices end up at originally red positions is $\frac{m}{n}$. What is $m+n$?
        
        \textit{\textbf{Mathematical Context:} The problem requires finding subsets $B$ such that $B \cap (B+k) = \emptyset$. Subsets of size $|B| \le 3$ trivially work. For $|B|=4$, exact inclusion-exclusion over rotational symmetries ($S_1, S_2, S_4, S_6$) is required to avoid overcounting invariant subsets.}
        
        \rule{\textwidth}{0.4pt}
        
        \begin{minipage}[t]{0.48\textwidth}
            \textbf{\textcolor{red}{Cons@512 (Baseline Failure Path)}} \\
            \textit{Model loses track of set intersections in long-chain enumeration.}
            
            \vspace{0.1cm}
            \textbf{Step 1:} 
            Correctly identifies that only subsets $|B| \le 4$ can satisfy the condition and correctly counts sizes 0 to 3 ($1+8+28+56 = 93$). \\
            \textbf{Step 2:} 
            For $|B|=4$, it groups invariant subsets by rotational periods but fails to properly intersect $S_2$ and $S_6$ with $S_4$. \\
            \textbf{Step 3:} 
            Erroneously concludes there are 24 valid subsets of size 4. $N = 93 + 24 = 117$. \\
            \textbf{Step 4:} 
            Blind scaling converges on 117/256 because the flawed inclusion-exclusion step is mechanically plausible.
            
            \vspace{0.1cm}
            \textbf{Final Answer:} \boxed{373} \quad \textcolor{red}{\textbf{(Incorrect)}}
        \end{minipage}
        \hfill
        \vline
        \hfill
        \begin{minipage}[t]{0.48\textwidth}
            \textbf{\textcolor{green!40!black}{\method{} (Ours --- Correct Path)}} \\
            \textit{Lookahead mechanism resolves combinatorial overlaps.}
            
            \vspace{0.1cm}
            \textbf{Step 1:} 
            Follows the same path to $|B|=4$. Confidence drops when attempting to sum the subsets $2+6+16$, sensing potential double-counting. \\
            \textbf{Step 2:} 
            Lookahead rollouts simulate specific subsets (e.g., $\{0,2,4,6\}$). The model detects that subsets in $S_1/S_3/S_5/S_7$ are entirely subsumed by $S_2/S_6$. \\
            \textbf{Step 3:} 
            Prunes the duplicate subsets, correcting the size 4 count from 24 to 22. \\
            \textbf{Step 4:} 
            $N = 93 + 22 = 115$. $m+n = 115 + 256 = 371$.
            
            \vspace{0.1cm}
            \textbf{Final Answer:} \boxed{371} \quad \textcolor{green!40!black}{\textbf{(Correct)}}
        \end{minipage}
    \end{tcolorbox}
    \caption{\textbf{Illustrative summary: \method{} concentrates compute on fragile enumeration steps.} In this AIME 2024 counting example, \texttt{Cons@512} preserves a mechanically plausible inclusion-exclusion error and overcounts the rotational cases, giving 373. \method{} identifies the low-confidence overlap calculation as the decision point, uses lookahead rollouts to reveal duplicated symmetry classes, and prunes the overcounted branch to recover the correct answer 371.}
    \label{fig:aime_case_study}
\end{figure}

\begin{figure}[!t]
    \centering
    \footnotesize
    \begin{tcolorbox}[colback=gray!5!white, colframe=gray!75!black, title=\textbf{Coherent Hallucination on GPQA} --- \textit{Illustrative Summary}]
        \small
        \textbf{Problem (GPQA Physics):} Ratio of solar-neutrino fluxes in Band 1 (700--800\,keV) vs.\ Band 2 (800--900\,keV). \textit{Context: the dominant $^7$Be line lies at 862\,keV, so the correct ratio is $\ll 1$.}

        \rule{\textwidth}{0.4pt}

        \textbf{Hallucination:} the trace maps $^7$Be neutrinos to $\approx$700\,keV $\rightarrow$ Band 1. \\
        \textbf{Reasoning:} Band 1 flux then dominates Band 2, so the ratio comes out inverted. \\
        \textbf{\method{} decision:} \textcolor{red}{\textbf{high ALC}} --- the continuation is internally consistent \emph{relative to the hallucinated fact}, so lookahead has no signal to veto it.
    \end{tcolorbox}
    \caption{\textbf{Illustrative summary: the knowledge-boundary failure mode of \S\ref{sec:gpqa_boundary}.} A hallucinated physical fact yields a coherent continuation that scores high ALC; internal confidence localizes instability, not falsehood.}
    \label{fig:neutrino_failure_compact}
\end{figure}

\begin{figure}[!t]
    \centering
    \begin{tcolorbox}[colback=gray!5!white, colframe=gray!75!black, title=\textbf{The Extraneous Root Trap in Continuous Mathematics} --- \textit{Illustrative Summary}]
        \small
        \textbf{Problem (Algebra / Equation Solving):} 
        Solve for real $x$: $\sqrt{x+2} - \sqrt{x-3} = 3$.
        
        \textit{\textbf{Mathematical Context:} Standard algebraic manipulation (squaring both sides) inherently introduces extraneous roots. A globally consistent solution must verify the derived roots against the original domain constraints ($x \ge 3$) and the equation itself.}
        
        \rule{\textwidth}{0.4pt}
        
        \begin{minipage}[t]{0.48\textwidth}
            \textbf{\textcolor{red}{Cons@512 (Baseline Failure Path)}} \\
            \textit{Model blindly executes local algebraic operations.}
            
            \vspace{0.1cm}
            \textbf{Step 1:} 
            Squares both sides: $(x+2) - 2\sqrt{(x+2)(x-3)} + (x-3) = 9$. \\
            \textbf{Step 2:} 
            Simplifies to $2x - 10 = 2\sqrt{x^2-x-6}$. Squares again to isolate $x$. \\
            \textbf{Step 3:} 
            Solves the resulting linear equation: $36x = 124 \implies x = \frac{31}{9}$. \\
            \textbf{Step 4:} 
            The algebraic steps are mechanically flawless. The model outputs the root without checking boundaries.
            
            \vspace{0.1cm}
            \textbf{Final Answer:} \boxed{\frac{31}{9}} \quad \textcolor{red}{\textbf{(Incorrect)}}
        \end{minipage}
        \hfill
        \vline
        \hfill
        \begin{minipage}[t]{0.48\textwidth}
            \textbf{\textcolor{green!40!black}{\method{} (Ours --- Correct Path)}} \\
            \textit{Lookahead mechanism anticipates domain violations.}
            
            \vspace{0.1cm}
            \textbf{Step 1:} 
            Squares both sides, mirroring the baseline. \\
            \textbf{Step 2:} 
            Confidence drops during the second squaring due to messy coefficients. The Lookahead rollout attempts to verify $x = \frac{31}{9} \approx 3.44$. \\
            \textbf{Step 3:} 
            Rollout detects that $\sqrt{3.44+2} - \sqrt{3.44-3} < 3$. ALC drops heavily for the squaring branch. \\
            \textbf{Step 4:} 
            \method{} prunes the branch and shifts to bounding analysis, realizing $\sqrt{x+2} - \sqrt{x-3}$ strictly decreases, yielding no real solutions.
            
            \vspace{0.1cm}
            {\raggedright\textbf{Final Answer:} \boxed{\text{Empty Set}} \textcolor{green!40!black}{\textbf{(Correct)}}\par}
        \end{minipage}
    \end{tcolorbox}
    \caption{\textbf{Illustrative summary: \method{} rejects deceptive algebraic branches through lookahead verification.} In this continuous-math example, \texttt{Cons@512} treats the result of repeated squaring as reliable and outputs an extraneous root. \method{} instead evaluates the downstream stability of that branch, checks the candidate against the original equation and domain behavior, and switches to the globally valid conclusion that no real solution exists.}
    \label{fig:algebra_case_study}
\end{figure}

\end{document}